\documentclass[lettersize,journal]{IEEEtran}
\usepackage{amsmath,amsfonts}
\usepackage{array}
\usepackage[caption=false,font=normalsize,labelfont=sf,textfont=sf]{subfig}
\usepackage{textcomp}
\usepackage{stfloats}
\usepackage{url}
\usepackage{verbatim}
\usepackage{graphicx}
\usepackage{cite}
\usepackage{hyperref}  
\usepackage[capitalise]{cleveref}

\usepackage{ifsym}
\usepackage{graphicx}
\usepackage{multirow}
\usepackage{utfsym}
\usepackage{bbding}
\usepackage{booktabs}
\usepackage[table,xcdraw]{xcolor}
\usepackage{float}
\definecolor{darkgreen}{RGB}{0,161,85}
\definecolor{darkred}{RGB}{255,0,58}
\usepackage{verbatim}

\definecolor{mygreen1}{RGB}{25,141,46} 
\definecolor{mybrown1}{RGB}{237,125,49} 
\definecolor{myblue1}{RGB}{0,112,192} 
\definecolor{myred1}{RGB}{191,70,51} 

\crefname{figure}{Fig.}{Figs.}
\Crefname{figure}{Fig.}{Figs.}
\crefname{table}{Table}{Tables}
\Crefname{table}{Table}{Tables}
\crefname{equation}{Eq.}{Eqs.}
\Crefname{equation}{Eq.}{Eqs.}
\crefname{section}{Section}{Sections}
\Crefname{section}{Section}{Sections}

\usepackage{ifsym}
\usepackage{graphicx}
\usepackage{multirow}
\usepackage{utfsym}
\usepackage{bbding}
\usepackage{booktabs}
\usepackage{float}
\usepackage{xcolor}
\definecolor{darkgreen}{RGB}{0,161,85}
\definecolor{darkred}{RGB}{255,0,58}
\usepackage{verbatim}

\usepackage{amsmath}
\usepackage{amsfonts}
\usepackage{amssymb}
\usepackage{wrapfig}
\usepackage{subcaption}
\usepackage{multirow}
 \usepackage{mathtools} 

\usepackage{verbatim}

\usepackage{anyfontsize}

\usepackage{microtype}
\usepackage{graphicx}
\usepackage{booktabs} 
\usepackage{multirow}
\usepackage{amsmath,amssymb}
\usepackage{booktabs}
\usepackage{caption,subcaption}

\usepackage{xcolor}
\definecolor{mygreen}{HTML}{3cb44b}
\definecolor{skyblue}{HTML}{beffff}
\definecolor{lightgreen}{HTML}{90ee90}

\usepackage{color, colortbl}

\definecolor{emerald}{rgb}{0.31, 0.78, 0.37}

\usepackage{tcolorbox}
\usepackage{enumitem}
\setitemize{itemsep=10pt,topsep=0pt,parsep=0pt,partopsep=0pt}
\usepackage{colortbl}

\usepackage{xcolor}
\definecolor{mygreen}{HTML}{3cb44b}
\colorlet{myyellow}{green!10!orange!90!}
\makeatletter

\usepackage{tikz}
\usetikzlibrary{arrows,shapes,snakes,automata,backgrounds,fit,petri}
\usepackage{adjustbox}

\newcommand{\RN}[1]{%
	\textup{\lowercase\expandafter{\it \romannumeral#1}}%
}
\usepackage{tabu}

\newcommand{\beq}{\vspace{0mm}\begin{equation}}
\newcommand{\eeq}{\vspace{0mm}\end{equation}}
\newcommand{\beqs}{\vspace{0mm}\begin{eqnarray}}
\newcommand{\eeqs}{\vspace{0mm}\end{eqnarray}}
\newcommand{\barr}{\begin{array}}
\newcommand{\earr}{\end{array}}

\usepackage{color, colortbl}
\definecolor{Gray}{gray}{0.93}

\usepackage{lipsum}

\usepackage{pifont}

\usepackage{makecell}

\usepackage{xcolor,amsmath}
\usepackage[linesnumbered,ruled,vlined]{algorithm2e}
\DontPrintSemicolon

\usepackage{xcolor}
\definecolor{mygreen}{HTML}{3cb44b}

\SetKwComment{Comment}{\color{green!50!black}\# }{}

\SetKwProg{Function}{def}{:}{}

\SetKwProg{For}{for}{:}{}
\SetKwProg{If}{if}{:}{}

\begin{document}
\title{REF-VLM: Triplet-Based Referring Paradigm for Unified Visual Decoding}

\author{
Yan~Tai,
Luhao~Zhu,
Yunan~Ding,
Yiying~Dong,
Guangtao~Zhai,~\IEEEmembership{Fellow,~IEEE},
Xiaohong~Liu,~\IEEEmembership{Member,~IEEE},
and Guodong~Guo,~\IEEEmembership{Senior~Member,~IEEE}
\thanks{Yan Tai is with the School of Computer Science, Shanghai Jiao Tong University, Shanghai, 200240, China, and also with the Ningbo Institute of Digital Twin, Eastern Institute of Technology, Ningbo, China.}
\thanks{Luhao Zhu, Yunan Ding, Yiying Dong, and Guodong Guo are with the Ningbo Institute of Digital Twin, Eastern Institute of Technology, Ningbo, China (gdguo@eitech.edu.cn).}
\thanks{Guangtao Zhai is with the School of Information Science and Electronic Engineering, Shanghai Jiao Tong University, Shanghai, 200240, China (email: zhaiguangtao@sjtu.edu.cn).}
\thanks{Xiaohong Liu is with the School of Computer Science, Shanghai Jiao TongUniversity, Shanghai, 200240, China (email: xiaohongliu@sjtu.edu.cn).}
\thanks{Corresponding Author: Xiaohong Liu, Guodong Guo.}
\thanks{The work was supported in part by the National Natural Science Foundation of China under Grant 62301310 and 62572317.}
}

\maketitle

\begin{abstract}
Multimodal Large Language Models (MLLMs) demonstrate robust zero-shot capabilities across diverse vision-language tasks after training on mega-scale datasets. However, dense prediction tasks, such as semantic segmentation and keypoint detection, pose significant challenges for MLLMs when represented solely as text outputs. Simultaneously, current MLLMs utilizing latent embeddings for visual task decoding generally demonstrate limited adaptability to both multi-task learning and multi-granularity scenarios. In this work, we present \textbf{REF-VLM}, an end-to-end framework for unified training of various visual decoding tasks. To address complex visual decoding scenarios, we introduce the \textbf{Triplet-Based Referring Paradigm (TRP)}, which explicitly decouples three critical dimensions in visual decoding tasks through a triplet structure: concepts, decoding types, and targets. TRP employs symbolic delimiters to enforce structured representation learning, enhancing the parsability and interpretability of model outputs. Additionally, we construct \textbf{Visual-Task Instruction Following Dataset (VT-Instruct)}, a large-scale multi-task dataset containing over 100 million multimodal dialogue samples across 25 task types. Beyond text inputs and outputs, VT-Instruct incorporates various visual prompts such as point, box, scribble, and mask, and generates outputs composed of text and visual units like box, keypoint, depth and mask. The combination of different visual prompts and visual units generates a wide variety of task types, expanding the applicability of REF-VLM significantly. Both qualitative and quantitative experiments demonstrate that our REF-VLM outperforms other MLLMs across a variety of standard benchmarks. The code, dataset, and demo will be publicly available. 
\end{abstract}

\begin{IEEEkeywords}
Multimodal Large Language Model, Computer Vision in the Wild, Multi-modality Learning
\end{IEEEkeywords}

\section{Introduction}
\label{sec:intro}

\IEEEPARstart{M}{ultimodal} Large Language Models (MLLMs) show strong performance in visual question answering and scene understanding \cite{llava, alayrac2022flamingo, lcl, 11261902}, yet their text-centric design limits fine-grained visual units decoding, hindering practical deployment in domains such as autonomous driving, robotics, and medical diagnosis.

\begin{figure*}[t]
    \centering
    \includegraphics[width=1.0\linewidth]{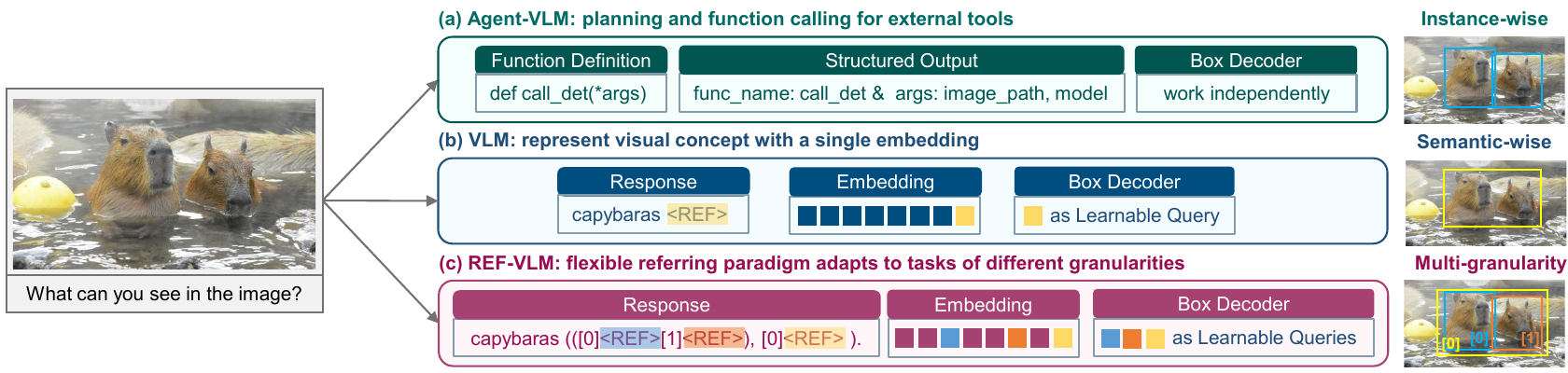}
    \caption{\textbf{Comparison of Visual Unit Decoding Methods.} Benefiting from the Triplet-Based Referring Paradigm, REF-VLM can adapt to more complex granularity scenarios and visual decoding tasks, enhancing the interpretability and accuracy of the MLLM's responses.}
    \label{fig:compare-ref}
\end{figure*}

As illustrated in \Cref{fig:compare-ref}(b), existing vision–language models (VLMs) \cite{rasheed2024glamm, lai2024lisa,visionllmv2} typically decode visual units (e.g., bounding boxes and masks) using a simplistic “\texttt{Visual Concept} + \texttt{Referring Token}” format, which is insufficient for complex, multi-granularity scenarios. To tackle more challenging visual decoding tasks (\Cref{fig:compare-ref}(c)), we propose the \textbf{Triplet-Based Referring Paradigm (TRP)}, which explicitly generates three components—(1) visual concept, (2) decoding type, and (3) referring tokens—via a structured special-token framework. Owing to its compositional triplet design, TRP provides a “one-fits-all” capability for unified multi-task and multi-granularity visual decoding.

To enhance the diversity of vision–language tasks, we introduce the \textbf{Visual-Task Instruction Following Dataset (VT-Instruct)}, a large-scale multimodal dataset designed to support a broad range of tasks, including visual understanding \cite{llava}, referring expressions \cite{10637468,chen2023shikra,you2023ferret,zhang2023next,zhang2024psalm}, interactive grounding (IG) \cite{zhang2024psalm,visionllmv2}, open-vocabulary identification \cite{10664543,visionllmv2,zhang2024psalm,visual_chat_gpt,shen2024hugginggpt}, grounded conversational generation (GCG) \cite{rasheed2024glamm}, keypoint detection \cite{visionllmv2,lin2023sphinx}, and depth estimation \cite{lin2023sphinx}. VT-Instruct contains over 100M high-quality multimodal dialogue samples curated primarily from publicly available datasets such as LAION-5B \cite{schuhmann2022laion}, SA-1B \cite{kirillov2023segment}, COCO \cite{lin2014microsoft}, and GRIT \cite{peng2023kosmos}. Each sample is augmented with carefully designed prompt templates involving multimodal inputs (e.g., images, text, points, boxes, scribbles, and masks) to support instruction following and diverse task-specific outputs.

Building upon the proposed decoding paradigm and the constructed dataset, we present \textbf{REF-VLM}, an end-to-end framework that enables unified multi-task training, contrasting with existing models that require separate stage-wise training for different tasks, thereby enhancing semantic consistency. To encode diverse user interactions, we introduce a novel \textit{parameter-free Mask-Guided Aggregation} scheme. Additionally, we propose a \textit{Latent Embeddings Router} and \textit{Parallel Grouped Hungarian Matching} to to enable the integration of multiple visual unit decoders in a joint multi-task training framework. Moreover, its plug-and-play architecture provides enhanced extensibility.

Our contributions can be summarized as follows:

\begin{itemize}
\setlength{\itemsep}{1pt}
\setlength{\parsep}{1pt}
\setlength{\parskip}{1pt}

\item We introduce a \textbf{Triplet-Based Referring Paradigm (TRP)} for precise referring under multi-task and multi-granularity settings. To support large-scale training under this paradigm, we construct \textbf{VT-Instruct}, a large-scale multimodal dataset comprising 100M dialogue samples across 25 visual task types, enabling robust understanding and decoding of diverse visual units.
\item We propose \textbf{REF-VLM}, an end-to-end framework for unified visual decoding, integrating novel components such as Mask-Guided Aggregation, Latent Embeddings Router, and Parallel Grouped Hungarian Matching to improve multi-task performance and adaptability.
\item Extensive experiments show that REF-VLM consistently outperforms existing MLLMs across diverse tasks, including visual understanding, referring expression comprehension, grounded conversational generation, open-vocabulary identification, and interactive grounding.

\end{itemize}

\section{Related Works}
\label{sec:related}

MLLMs often lack the capability to output visual units such as boxes, keypoints, and masks. To expand their applicability in real-world visual tasks, it is typically necessary to implement targeted designs for different visual tasks.

\noindent \textbf{Decode Visual Units as Sequences.} The most straightforward solution leverages the text generation capabilities of MLLM to produce visual localization results in textual format \cite{chen2023shikra, chen2021pix2seq, peng2023kosmos, ma2024groma}. This approach does not require structural modifications to the MLLM. Pix2Seq \cite{chen2021pix2seq} leverages the model's autoregressive generation capability to express bounding boxes and class labels as sequences of discrete tokens. Shikra \cite{chen2023shikra} constructs an visual supervised fine-tuning dataset, where the model needs to perform inductive analysis before answering complex questions, and outputs bounding boxes in text form.

\noindent \textbf{Decode Visual Units with Agent Tools.} Another approach involves using the MLLM as an agent to coordinate task-specific models, enabling localization of visual targets \cite{llava-plus}. In this case, MLLM outputs textual descriptions of recognized content and scheduling results, which can be utilized by downstream visual tools. LLaVa-Plus \cite{llava-plus} constructs an instruction-following dataset that includes a large number of samples for using task-specific models as tools. VITRON \cite{fei2024vitron} incorporates a sketch encoder to process user-provided visual prompts and supports direct generation of bounding box coordinates by the model. However, since the final visual units is derived from the tool models, there may be a gap between the MLLM's understanding and the final output. Moreover, the model is unable to effectively leverage previous visual recognition results during prediction, requiring repeated input of tool model outputs, leading to issues such as insufficient robustness and accuracy in multi-task and multi-target applications.

\noindent \textbf{Decode Visual Units with Latent Embeddings.} Using the tokens output by the MLLM as learnable queries input into task-specific decoders is the most widely adopted visual decoding strategy \cite{rasheed2024glamm, lai2024lisa}. LISA \cite{lai2024lisa} adopts SAM \cite{kirillov2023segment} as the mask decoder, where MLLM generates learnable special tokens as prompts for SAM, producing fine-grained segmentation results. PSALM \cite{zhang2024psalm} divides the input for open-vocabulary segmentation tasks into instruction prompts, condition prompts, and discrete mask tokens, decoding the output mask tokens to obtain segmentation results aligned with the prompt content. GLAMM \cite{rasheed2024glamm} integrates text-based dialogue with segmentation tasks, utilizing SAM to simultaneously generate detailed descriptions and mask results from the model. VisionLLM v2 \cite{visionllmv2} introduces the super-link technique, where the MLLM generates task-specific special tokens to serve as routing tokens. These tokens are followed by additional learnable queries appended to them, facilitating visual decoding tasks.

REF-VLM adopts a latent embedding-based approach. Unlike VisionLLM v2~\cite{visionllmv2}, which relies on task-specific special tokens and therefore often necessitates retraining the LLM when adapting to new decoders, REF-VLM utilizes a unified \texttt{<REF>} token to represent diverse visual units. This design leverages the LLM’s language capabilities to distinguish tasks without additional training. 

\begin{figure*}[!htbp]
  \centering
  \includegraphics[width=1.0\linewidth]{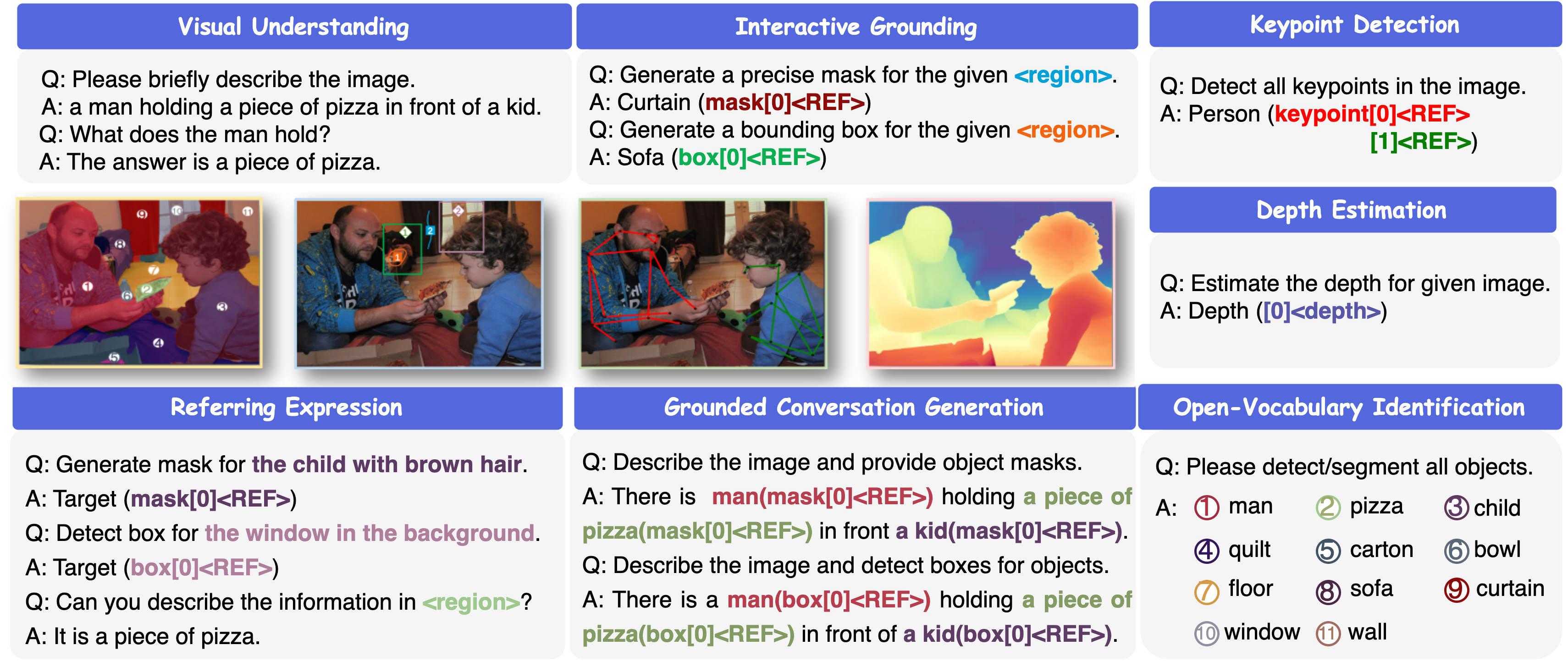}
  \caption{\textbf{Example of VT-Instruct Dataset by Using the Automated Data Construction Pipeline.} Our VT-Instruct dataset contains seven distinct downstream tasks, including Visual Understanding, Referring Expression, Interactive Grounding, Grounded Conversation Generation, Open-Vocabulary Identification and Depth Estimation.}
  \label{fig:data_example}
\end{figure*}

\section{Unified Instruction Pipeline}
\label{sec:pipeline}

\subsection{Triplet-Based Referring Paradigm}\label{sec:trp}

As outlined in the preceding section, two dense prediction decoding approaches are introduced: ``Agent Tools" and ``Latent Embeddings". As shown in \Cref{fig:compare-ref}(a), MLLMs act as agents, generating structured text (e.g., JSON) to invoke external decoders like Grounding DINO \cite{liu2023grounding} and SAM \cite{kirillov2023segment}. Meanwhile, \Cref{fig:compare-ref}(b) and (c) depict the Latent Embeddings approach, where MLLM outputs serve as learnable queries for visual task decoders. Here, the special token \texttt{<REF>} facilitates referential learning. Our REF-VLM adopts the Latent Embeddings framework, offering greater flexibility and adaptability for diverse visual decoding tasks. For clarity, we omit some special tokens used for assisting referential tasks.

In \Cref{fig:compare-ref}(b), we illustrate the conventional referring paradigm \cite{visionllmv2, rasheed2024glamm, lai2024lisa, ren2024pixellm}, where the \texttt{<REF>} token is typically introduced after visual concepts to enable a single decoding process. However, discrete labels suffer from semantic ambiguity. For instance, the phrase ``\texttt{People are crossing the street}" can refer to visual concepts such as the entire scene (single target), the people (multiple targets) and the street (single target). Existing reference schemes struggle to effectively handle visual concept references at different granularities, ultimately impacting the interpretability and accuracy of MLLM responses. Moreover, since different visual tasks require varying decoding granularities, extending the ``Latent Embeddings" decoding approach to multi-task scenarios necessitates a more effective referring and embedding framework.

We propose the \textbf{Triplet-Based Referring Paradigm (TRP)}, a mechanism for multi-granularity visual concept decoding. As shown in \Cref{fig:compare-ref}(c), TRP resolves semantic ambiguity and supports multi-granularity referencing in visual tasks. TRP comprises three components: \textit{(i) Visual Concepts}, encapsulated in \texttt{<Phrase>} tags (e.g., \texttt{<Phrase>dogs</Phrase>}); \textit{(ii) Decoding Types}, specified in \texttt{<Unit>} tags (e.g., \texttt{<Unit>box</Unit>}); and \textit{(iii) References}, denoted by \texttt{<REF>} to link concepts to instances (e.g., \texttt{[0]<REF>} for the first detected dog). Consequently, the example in \Cref{fig:compare-ref}(c) corresponds to the full representation: \texttt{<Phrase>capybaras</Phrase> ((<Unit>box</Unit>[0]<REF>[1]<REF>),<Unit> box</Unit>[0]<REF>)}. 

The proposed TRP demonstrates a ``one-fits-all" advantage across diverse visual tasks. This is achieved through two key design strengths: (1) Syntactic Scalability: The triplet-based structure inherently supports compositional expansion. TRP can represent complex scene descriptions through hierarchical nesting, such as ``\texttt{<Phrase><Phrase>dog</Phrase> in <Phrase>park</Phrase></Phrase>}". Moreover, TRP can specify composite tasks by combining multiple ``\texttt{<Unit>}" tags (e.g., ``\texttt{<Unit>box, keypoint</Unit>}"). (2) Task Extensibility: By predefining the semantic space of ``\texttt{<Unit>}" tags (e.g., introducing ``\texttt{<Unit>depth</Unit>}"), TRP enables zero-shot task extension, allowing seamless adaptation to new visual tasks without additional training \cite{lcl}. These design strengths ensure that TRP is not only versatile but also future-proof, making it a robust solution for multi-granularity visual concept decoding. Unlike VisionLLM v2 \cite{visionllmv2}, which uses special tokens as routing embeddings and introduces predefined tokens like ``\texttt{[DET]}" and ``\texttt{[SEG]}" as routing tokens, TRP adopts a more flexible multi-task instruction tuning approach. In this scheme, the MLLM itself serves as the router, deciding the decoding type and representing it in text (denoted as \texttt{Unit}). When the model faces incremental task requirements, the routing token approach necessitates an expansion of the MLLM's vocabulary. In contrast, TRP can leverage the generative capabilities of large models, allowing for low-cost adaptation.

\begin{table*}[!htbp]
\centering
\caption{\textbf{Data statistics of VT-Instruct and actual use of dataset in the training process.} Multiple datasets were utilized to train REF-VLM, with some supporting multiple tasks. For instance, GranD \cite{rasheed2024glamm} enables tasks such as captioning, REC, RES, REG, GCG, and open-vocabulary identification. Most datasets within VT-Instruct were employed as subsets in our training process.}
\label{tab:data_list}
\resizebox{\textwidth}{!}{%
\begin{tabular}{c|c|c c c}
\hline
\rowcolor[HTML]{CCCCCC} 
Task & Sub-Task & Original Dataset & Construction Number & Actual Use \\ \hline
\multirow{2}{*}{\begin{tabular}[c]{@{}c@{}}Visual \\ Understanding\end{tabular}} 
 & Caption & COCO \cite{lin2014microsoft}, GranD \cite{rasheed2024glamm}, GRIT \cite{peng2023kosmos}& 15,980,000 & 780,000 \\ 
 & VQA & VQAv2 \cite{VQA}, LLaVA-Instruct \cite{llava}, VCR \cite{zellers2019vcr} & 1,310,000 & 1,310,000 \\ \hline
\multirow{3}{*}{\begin{tabular}[c]{@{}c@{}}Referring \\ Expression\end{tabular}} 
 & REC & \begin{tabular}[c]{@{}c@{}}RefCOCO \cite{kazemzadeh2014referitgame}, RefCOCO+ \cite{kazemzadeh2014referitgame}, RefCOCOg \cite{kazemzadeh2014referitgame},\\ GranD \cite{rasheed2024glamm}, GRIT \cite{peng2023kosmos}\end{tabular} & 22,880,000 & 880,000 \\ 
 & RES & \begin{tabular}[c]{@{}c@{}}RefCOCO \cite{kazemzadeh2014referitgame}, RefCOCO+ \cite{kazemzadeh2014referitgame}, RefCOCOg \cite{kazemzadeh2014referitgame},\\ GranD \cite{rasheed2024glamm}\end{tabular} & 3,880,000 & 680,000 \\ 
 & REG & \begin{tabular}[c]{@{}c@{}}RefCOCO \cite{kazemzadeh2014referitgame}, RefCOCO+ \cite{kazemzadeh2014referitgame}, RefCOCOg \cite{kazemzadeh2014referitgame},\\ GranD \cite{rasheed2024glamm}, GRIT \cite{peng2023kosmos}, COCO-Interactive \cite{zhang2024psalm},\\ Osprey \cite{yuan2024osprey}, Visual Genome \cite{krishna2017visual}, Visual7W \cite{zhu2016visual7w}\end{tabular} & 22,750,000 & 1,200,000 \\ \hline
\multirow{3}{*}{\begin{tabular}[c]{@{}c@{}}Interactive \\ Grounding\end{tabular}} 
 & IG-Box & COCO-Interactive \cite{zhang2024psalm} & 3,200,000 & 120,000 \\ 
 & IG-Mask & COCO-Interactive \cite{zhang2024psalm} & 3,200,000 & 120,000 \\ 
 & IG-Keypoint & COCO \cite{lin2014microsoft} & 500,000 & 140,000 \\ \hline
\multirow{2}{*}{\begin{tabular}[c]{@{}c@{}}Grounded \\ Conversation \\ Generation\end{tabular}} 
 & GCG-box & GRIT \cite{peng2023kosmos}, GranD \cite{rasheed2024glamm}, Flickr30k-Entities \cite{plummer2015flickr30k}& 15,630,000 & 540,000 \\ 
 & GCG-mask & \begin{tabular}[c]{@{}c@{}}GranD \cite{rasheed2024glamm}, LLaVA-Grounding \cite{zhang2023llava} , PNG \cite{gonzalez2021panoptic},\\ OpenPSG \cite{zhou2024openpsg}\end{tabular} & 4,000,000 & 450,000 \\ \hline
\multirow{2}{*}{\begin{tabular}[c]{@{}c@{}}Open-Vocabulary \\ Identification\end{tabular}} 
 & OVD/FOVD & GranD \cite{rasheed2024glamm}, GRIT \cite{peng2023kosmos}, COCO-REM \cite{cocorem} & 15,770,000 & 600,000 \\ 
 & OVS/FOVS & \begin{tabular}[c]{@{}c@{}}GranD \cite{rasheed2024glamm}, COCO-REM \cite{cocorem}, ADE20k \cite{zhou2017scene},\\ Cityscapes \cite{cordts2016cityscapes}\end{tabular} & 3,795,000 & 600,000 \\ \hline
Keypoint Detection & - & COCO \cite{lin2014microsoft} & 140,000 & - \\ \hline
Depth Estimation & - & Kitti \cite{kitti} , HRWSI \cite{hrwsi}, NYU \cite{nyu} & 150,000 & - \\ \hline
\end{tabular}%
}
\end{table*}

\begin{table*}[h]
    \centering
    \caption{\textbf{Prompt template for evaluating different kind of tasks.} For different evaluation tasks, we utilized distinct prompt templates. During the actual training process, to ensure the model's generalization across tasks, we constructed at least 100 templates for each subtask.}
    \label{tab:eval_tempalte}
    \resizebox{\textwidth}{!}{%
    \begin{tabular}{c|l}
    \hline
    \rowcolor[HTML]{CCCCCC} 
    Task & \multicolumn{1}{c}{Template}\\ \hline
    Caption & \textless image\textgreater Please describe the image in detail. \\
    VQA & Please take a look at the image \textless image\textgreater and promptly provide an answer for \textless question\textgreater. \\
    GCG-Mask & Describe the setting of the image \textless image\textgreater and offer masks for each visible object. \\
    GCG-Box & Please describe the image \textless image\textgreater  and detect relevant bounding boxes. \\
    REC & What are the coordinates of \textless referring expression\textgreater in the image\textless image\textgreater? \\
    RES & Provide a segmentation mask for \textless referring expression\textgreater in the picture \textless image\textgreater. \\
    REG & For the given image \textless image\textgreater, can you provide a unique description of the area \textless mask\textgreater? \\
    IG-Mask & Please generate a mask based on the region \textless region\textgreater in the image \textless image\textgreater. \\
    FOVD & Please detect bounding boxes in the image\textless image\textgreater. \\
    FOVS & Please segment objects in the image\textless image\textgreater. \\ 
    Keypoint Detection & Please detect all the people and visualize all the keypoints in the image\textless image\textgreater. \\
    \hline
    \end{tabular}%
    }
\end{table*}

\begin{figure*}[t]
  \centering
   \includegraphics[width=1.0\linewidth]{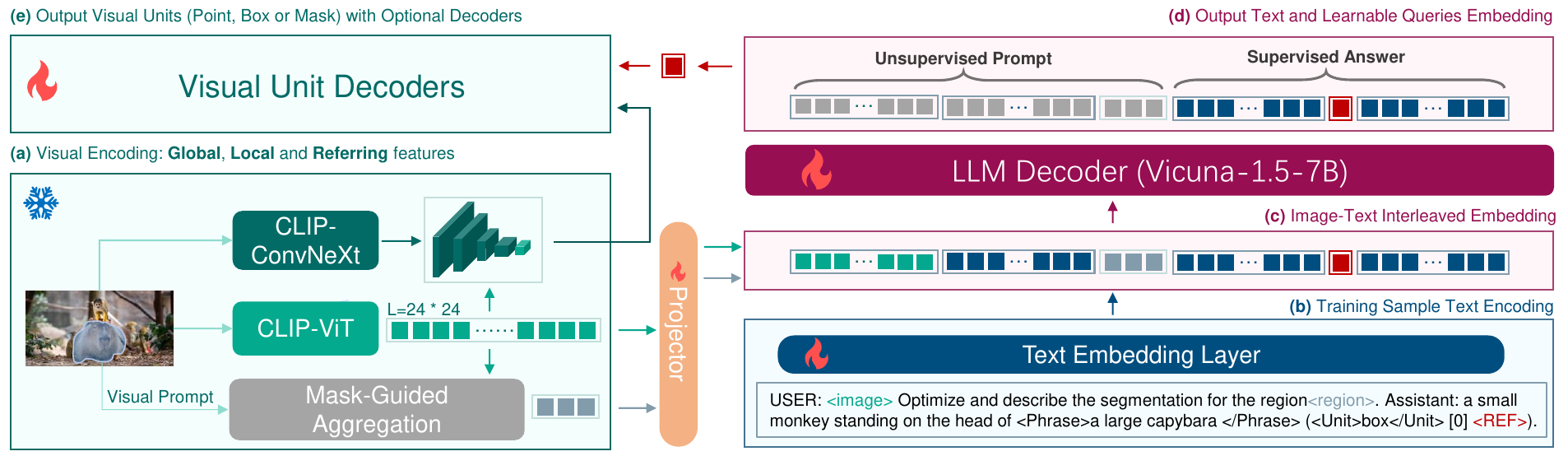}
  \caption{\textbf{The Framework of REF-VLM.} REF-VLM employs dual-architecture visual encoders to jointly encode images into a feature pyramid, enhancing visual unit decoder performance. Additionally, visual prompts are fused with global features and share a projector, enabling parameter-free encoding of image interactions. Training samples adhere to the Triplet-Based Referring Paradigm, ensuring one-to-one mapping between REF-VLM's latent embeddings and decoding targets.
    }
  \label{fig:framework}
\end{figure*}

\subsection{Visual-Task Instruction Following Dataset}
\label{sec:vt-instruct}

Following the Triplet-Based Referring Paradigm, We present \textbf{Visual-Task Instruction Following Dataset (VT-Instruct)}, a large-scale visual multi-task dataset that combines different visual prompts as inputs and visual units as outputs. VT-Instruct comprises over 100 million dialogue samples featuring multimodal input-output pairs. These pairs encompass various combinations of output units, ranging from low to high visual density, including point, box, keypoint, depth and mask, combined with either low or high text complexity. 

VT-Instruct includes 7 different downstream tasks, as shown in \Cref{fig:data_example}. The Visual Understanding task includes Image Captioning and Visual Question Answering (VQA), involving image-text inputs and text-only outputs. Referring Expression tasks cover Referring Expression Comprehension (REC), Referring Expression Segmentation (RES), and Referring Expression Generation (REG). While REC and RES require models to predict bounding boxes or masks in response to a query about a specific region in an image, REG involves generating descriptive text from visual inputs like points, boxes, scribbles, or masks. Interactive Grounding (IG) enables users to provide prompts via both text and interactive inputs (e.g., points, boxes, masks), allowing MLLMs to interpret and generate corresponding outputs. Open-Vocabulary Identification focuses on localizing and segmenting objects from descriptive text, even if the object categories were not part of the training data. For traditional Open-Vocabulary tasks performed by current MLLM, it typically requires user inputs specific class names, Grounded Conversation Generation (GCG) produces natural language responses interwoven with bounding boxes or masks, with the GCG task further divided into GCG-box (bounding box outputs) and GCG-mask (mask outputs). For each downstream task, we \textbf{(i)} first construct a specific system instruction and \textbf{(ii)} generate over 150 task-specific prompt templates using GPT-4, randomly selecting them to construct user prompts (see \Cref{tab:eval_tempalte}), then \textbf{(iii)} we modify existing dataset annotations to construct a unified answering format following the rule of TRP, creating multi-turn conversations featuring a system-prompt-answer combination. Specific, for each task, we select a unique prompt-unit pair to develop task-specific instructions. For example, visual understanding task encompasses Image Captioning and Visual Question Answering (VQA), with image-text inputs and pure text outputs. To facilitate MLLMs in comprehending image-level information and addressing diverse questions, we construct conversations for visual understanding tasks using our proposed pipeline with the COCO \cite{lin2014microsoft}, GranD \cite{rasheed2024glamm}, GRIT \cite{peng2023kosmos}, VQAv2 \cite{VQA}, and LLaVA-instruct \cite{llava} datasets, which collectively comprise over 15 million image-text pairs featuring multi-turn conversations. Referring expression tasks include Referring Expression Comprehension (REC), Referring Expression Segmentation (RES), and Referring Expression Generation (REG). The REC and RES tasks require the model to respond to a question or description regarding a specific area in an image, predicting bounding boxes or masks. In contrast, the REG task involves inputs such as points, boxes, scribbles, and masks, with the model expected to generate a descriptive response based on the visual prompts. We construct conversations for referring expression task from refCOCO \cite{kazemzadeh2014referitgame}, refCOCO+ \cite{kazemzadeh2014referitgame}, refCOCOg \cite{kazemzadeh2014referitgame}, GranD \cite{rasheed2024glamm}, GRIT \cite{lin2014microsoft}, Osprey \cite{yuan2024osprey}, Visual Genome \cite{krishna2017visual} datasets with more than 22 million samples. Interactive grounding allows users to provide prompts through both text and interactive elements, such as points, boxes, masks, or scribbles, enabling MLLMs to interpret these inputs and generate corresponding outputs, including bounding boxes or masks. We constructed interactive grounding samples using the COCO-interactive \cite{zhang2024psalm} dataset , which contains over 64 million examples. The open-vocabulary identification task focuses on localizing and segmenting objects in an image based on descriptive text prompts, even if the specific object categories were not included in the model's training data. To equip REF-VLM with zero-shot capabilities for object detection and segmentation—similar to traditional open-vocabulary detection models (e.g., YOLO-World \cite{cheng2024yolo}) and segmentation models (e.g., SAM \cite{kirillov2023segment}) — we designed a multimodal conversation system using bounding boxes and masks annotations from the GRIT \cite{peng2023kosmos}, GranD \cite{rasheed2024glamm}, COCO-REM \cite{cocorem}, ADE20k \cite{zhou2017scene}, and Cityscapes \cite{cordts2016cityscapes} datasets, resulting in a corpus of over 20 million examples. Grounded conversation generation (GCG) aims to produce natural language responses interwoven with bounding boxes or object segmentation masks. The GCG task is divided into GCG-box, which outputs bounding boxes, and GCG-mask, which outputs masks. We developed these tasks using datasets that include captions and phrases associated with bounding box or mask annotations, such as Flickr30k-entities \cite{plummer2015flickr30k}, GranD \cite{rasheed2024glamm}, GRIT \cite{peng2023kosmos}, LLaVA-grounding \cite{zhang2023llava}, OpenPSG \cite{zhou2024openpsg}, and PNG \cite{gonzalez2021panoptic}, collectively comprising over 18 million annotations. 

As shown in \Cref{tab:data_list}, VT-Instruct is constructed entirely from publicly available datasets without any in-house data. Due to efficiency and resource constraints, REF-VLM is trained on only a small subset of VT-Instruct. As a visual decoding task framework, REF-VLM’s training samples are generally comparable to or fewer than those in similar studies \cite{alayrac2022flamingo, visionllmv2, rasheed2024glamm, bai2023qwen, ren2024pixellm, zhang2024psalm, yuan2024osprey, chen2023internvl} for each individual task.

\section{End-to-End Decoding Framework}

Unlike existing ``Latent Embeddings" decoding methods that require task-specific fine-tuning in separate stages \cite{visionllmv2}, REF-VLM achieves \emph{unified end-to-end training} for all tasks, including conventional QA, VQA, and various visual decoding tasks. We will illustrate the training process using an example based on the Referring GCG-Segmentation task and discuss the core components of the framework in the following subsections.

\subsection{Unified Training Workflow}\label{sec:training-flow}

The example training task in \Cref{fig:framework} requires the MLLM to describe a user-specified region based on a mask and prompt input, and generate precise segmentation results.

REF-VLM supports image and visual prompt (VPT) inputs, where VPT includes point, box, scribble, and mask. For image encoding, we follow LLaVA \cite{llava} and use CLIP-ViT \cite{clip} to extract global features mapped to the text embedding space. To address CLIP-ViT's limitations \cite{rasheed2024glamm, lai2024lisa, zhang2024psalm} in dense prediction tasks, we additionally employ CLIP-ConvNeXt \cite{cherti2023reproducible}, a Conv-based architecture pre-trained on large-scale image-text pairs, to capture multi-scale local features. The outputs of both encoders are concatenated to improve visual task decoding. For VPT encoding, we propose a parameter-free \emph{Mask-Guided Aggregation} method to fuse image and VPT features, enabling precise region understanding.

REF-VLM employs Vicuna-v1.5-7B \cite{zheng2023vicuna} as the base LLM for processing the text modality. \Cref{fig:framework}(b), (c), and (d) illustrate the parallel decoding and supervision process of a training sample. In the \textit{(b) Training Sample Text Encoding} process, the input training text adheres to the TRP specification described earlier, requiring the LLM to generate a sequence of latent embeddings of equal length for the final visual decoding target, represented as \texttt{<REF>} tokens. In pipeline \textit{(c) Image-Text Interleaved Embedding}, the image global features and aggregated visual prompt features output from pipeline \textit{(a) Visual Encoding} are substituted into the reserved placeholders (i.e., \texttt{<image>} and \texttt{<region>}) respectively, forming a complete sequence fed into the LLM Decoder. 

\Cref{fig:framework}(d) and (e) illustrates the supervision process of the training sequence. The gray area corresponds to the prompt part, which does not require loss calculation during the supervised fine-tuning. The blue area corresponds to the answer part, which is mapped to the vocabulary and included in the loss computation. The red area corresponds to the \texttt{<REF>} tokens, which not only undergo the same LLM loss calculation as the blue area but also serve as inputs to the visual unit decoders for further decoding.

REF-VLM follows a two-stage training process. In the first stage, similar to Shikra \cite{chen2023shikra}, only the global visual encoder (CLIP-ViT), the projector, and the LLM participate in computation. During this phase, the weights of CLIP-ViT and the LLM remain fixed, with only the projector's parameters being updated. In the second stage, REF-VLM is trained in a unified manner across all tasks. Beyond the modules in the \textit{(a) Visual Encoding pipeline}, all other components, including the projector, LLM, and visual unit decoders are updated. The Unified Training Workflow of REF-VLM offers better semantic consistency compared to similar approaches that rely on pre-trained visual task decoders \cite{visionllmv2, rasheed2024glamm, ren2024pixellm, lai2024lisa, yuan2024osprey}, such as SAM \cite{kirillov2023segment} or Grounding-DINO \cite{liu2023grounding}. REF-VLM eliminates the need to repeatedly use visual encoders for each visual task, significantly reducing the overall model parameters.

\begin{figure*}[t]
    \centering
    \includegraphics[width=1.0\linewidth]{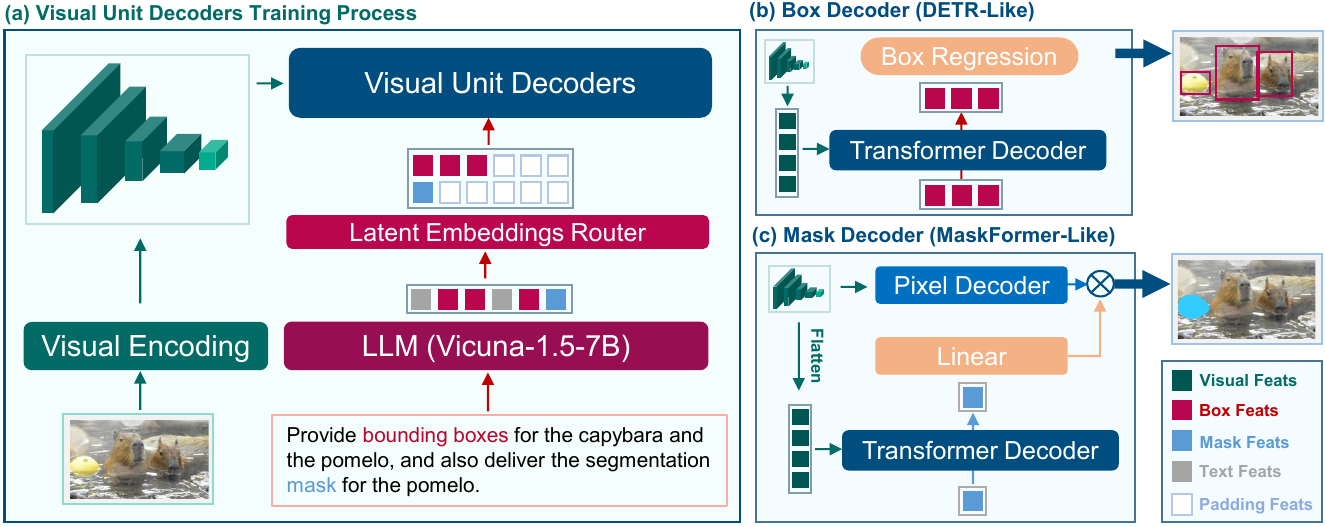}
    \caption{\textbf{Architecture of Visual Unit Decoders.} We propose a Latent Embeddings Router to facilitate unified multi-task training in REF-VLM, and enhance the Hungarian matching algorithm for the TRP-based one-to-one referring decoding scheme.}
    \label{fig:unit_decoders}
\end{figure*}

\subsection{Mask-Guided Aggregation}\label{sec:vpt-encoder}

REF-VLM achieves diverse referencing tasks through visual prompt support (points, boxes, scribbles, masks). Unlike existing methods requiring additional parameters \cite{yuan2024osprey, fei2024vitron}, we introduce \textbf{parameter-free Mask-Guided Aggregation}. First, we perform two steps: (1) converting prompts into normalized masks whose sizes are aligned with global features; (2) partitioning features and masks into grids for patch-wise fusion.

Let the input image global features be denoted as \( \mathcal{X} \in \mathbb{R}^{C \times N \times H \times W} \), where \( C \) is the number of channels, \( N \) indicates the count of spatial patches, and \( H \times W \) define the spatial resolution of each patch. Given a mask \( \mathcal{M} \in \mathbb{R}^{Q \times N \times H \times W} \) with \( Q \) representing the number of masks in the input prompt, our aggregation operation is implemented as an Einstein summation (tensor contraction) between the feature tensor \( \mathcal{X} \) and the query-specific mask \( \mathcal{M} \), contracting over the spatial dimensions \( (h, w) \).
\begin{equation}
\label{eq:matching}
    \mathcal{V}_{q,n,c} = \sum_{h=1}^{H}\sum_{w=1}^{W} \mathcal{X}_{c,n,h,w} \cdot \mathcal{M}_{q,n,h,w}.
\end{equation}
Let the aggregated output tensor be $\mathcal{V} \in \mathbb{R}^{Q \times N \times C}$. To inject spatial awareness, we augment it with cosine positional encodings. For position index $n$ and channel $c$, the encoding is defined as:  
\begin{equation}
\text{PE}(n, c) = \cos\left( \frac{n}{s^{2c/C}} \right),
\end{equation}  
where $s$ is a temperature hyperparameter. The refined features $\widetilde{\mathcal{V}}$ are obtained by:  
\begin{equation}
\widetilde{\mathcal{V}}_{q,n,c} = \mathcal{V}_{q,n,c} + \alpha \cdot \text{PE}(n, c),
\end{equation}  
where $\alpha$ could be a learnable scalar, but in our setup, $\alpha$ is set to a constant value of 1.

Finally, the visual prompt features and image global features share the projector layer, achieving efficient feature fusion without introducing additional parameters.

\subsection{Visual Unit Decoders}\label{sec:unit-decoder}

In unified multi-task training, each batch instance undergoes task-specific decoding processes. \Cref{fig:unit_decoders} illustrates REF-VLM's workflow, showing a single sample requiring dual-task decoding. \Cref{fig:unit_decoders}(a) shows REF-VLM's training pipeline: (1) visual data passes through dual encoders to build a feature pyramid; (2) the LLM processes prompts, generating TRP-structured responses with \texttt{<REF>} tokens for one-to-one referring between latent embeddings and visual instances. The LLM embeddings are categorized as: gray (text tokens), red (box targets embeddings), and blue (mask targets embeddings). We adopt streamlined yet effective decoders: DETR \cite{carion2020detr} for box prediction and MaskFormer \cite{cheng2021maskformer} for mask segmentation. As shown in \Cref{fig:unit_decoders}(a)(b), fused image features are flattened and fed into a transformer decoder alongside latent embeddings.

\noindent \textbf{Latent Embeddings Router.} To address synchronization issues caused by unused decoders during multi-task training, we introduce a Latent Embeddings Router. It extracts \texttt{<REF>} embeddings by decoding type, pads them to a fixed length $L=100$, and feeds them into corresponding decoders as inputs of shape $\mathbb{R}^{L \times d}$, where $d$ is the embedding dimension. When no targets exist for a decoder, its routed features still participate in forward computation, but their loss is set to zero to maintain training stability.

\noindent \textbf{Parallel Grouped Hungarian Matching.} In DETR-style \cite{carion2020detr} object detection, each learnable query is classified and matched to ground-truth targets using Hungarian matching. In contrast, REF-VLM uses TRP to ensure that each \texttt{<REF>} token is aligned with a unique referring phrase, removing the need for a classification head. For example, given an input like “\texttt{dogs [0]\texttt{<REF>}[1]\texttt{<REF>}, cats [0]\texttt{<REF>}[1]\texttt{<REF>}[2]\texttt{<REF>}},” the \texttt{<REF>} tokens are inherently grouped by phrase semantics (e.g., “dogs” vs. “cats”). During training, only tokens within the same group are matched, avoiding cross-entity interference. We propose Parallel Grouped Hungarian Matching to perform efficient, group-wise matching of model outputs, enabling scalable distributed training.

Let $\mathcal{B}$ denote a batch containing $B$ independent groups. For the $i$-th group: (i) $\mathbf{P}^{(i)} = \{\mathbf{p}_1^{(i)},...,\mathbf{p}_{N_i}^{(i)}\}$: Set of $N_i$ predicted boxes. (ii) $\mathbf{T}^{(i)} = \{\mathbf{t}_1^{(i)},...,\mathbf{t}_{M_i}^{(i)}\}$: Set of $M_i$ target boxes. (iii) $\mathcal{C}^{(i)} \in \mathbb{R}^{N_i \times M_i}$: Cost matrix where element $c_{nm}^{(i)}$ represents the matching cost between $\mathbf{p}_n^{(i)}$ and $\mathbf{t}_m^{(i)}$. (iv) $\sigma^{(i)}: \{1,...,N_i\} \rightarrow \{1,...,M_i\} \cup \{\varnothing\}$: Injective matching function for the $i$-th group. The pairwise cost combines geometric measures:
\begin{equation}
c_{nm}^{(i)} = \lambda_{\text{L1}} \mathcal{L}_{\text{L1}}(\mathbf{p}_n^{(i)}, \mathbf{t}_m^{(i)}) + \lambda_{\text{GIoU}} \mathcal{L}_{\text{GIoU}}(\mathbf{p}_n^{(i)}, \mathbf{t}_m^{(i)}),
\end{equation}
where $\lambda_{\text{L1}}$ and $\lambda_{\text{GIoU}}$ are weighting coefficients, $\mathcal{L}_{\text{L1}}$ denotes normalized coordinate differences, and $\mathcal{L}_{\text{GIoU}}$ represents the generalized IoU loss.
To enable parallel computation, we construct a padded tensor:
\begin{equation}
\widetilde{\mathcal{C}} \in \mathbb{R}^{B \times N_{\text{max}} \times M_{\text{max}}},
\end{equation}
where $N_{\text{max}} = \max(N_i)$ and $M_{\text{max}} = \max(M_i)$. Invalid entries are masked using:
\begin{equation}
\mathbf{M}^{(i)}_{nm} = \begin{cases}
0 & n \leq N_i \land m \leq M_i \\
-\infty & \text{otherwise}
\end{cases}.
\end{equation}
For each group $i$, find optimal permutation matrix $\widetilde{\mathbf{A}}^{(i)} \in \{0,1\}^{N_{\text{max}} \times M_{\text{max}}}$ that minimizes:
\begin{equation}
\sum_{n=1}^{N_{\text{max}}} \sum_{m=1}^{M_{\text{max}}} \widetilde{\mathcal{C}}^{(i)}_{nm} \widetilde{\mathbf{A}}^{(i)}_{nm},
\end{equation}
subject to:
\begin{equation}
\sum_{m=1}^{M_{\text{max}}} \widetilde{\mathbf{A}}^{(i)}_{nm} \leq 1 \quad \forall n, \quad \sum_{n=1}^{N_{\text{max}}} \widetilde{\mathbf{A}}^{(i)}_{nm} \leq 1 \quad \forall m.
\end{equation}

\subsection{Training Details}
\label{sec:train_details}

The training process of REF-VLM is conducted in three stages, during which both CLIP-ViT and CLIP-ConvNeXt are frozen, with no parameter updates. We use eight NVIDIA A800-80GB GPUs in all of our training processes and pick Vicuna-7B as our LLM, CLIP-large-14-336 and CLIP-ConvNeXt-512 as our visual encoder. 

\textbf{Stage 1.} In the first stage, we train the projector to equip the LLM with the ability to interpret visual information and establish effective cross-modal alignment. REF-VLM follows the same training setup as Shikra \cite{chen2023shikra}, where all model parameters are frozen except for the projector, allowing the training process to focus exclusively on bridging visual and textual representations. This stage mainly involves visual understanding tasks, such as image captioning and VQA, which serve as foundational supervision to improve the projector’s visual–text alignment capability. For object-centric tasks such as REC and GCG-box, no task-specific visual decoders are introduced at this stage. Instead, following \cite{chen2023shikra}, numerical bounding box coordinates are serialized as text strings and provided as part of the model input, enabling the LLM to acquire a coarse sense of spatial relationships without explicit visual decoding. This design allows the model to build basic spatial reasoning ability while keeping the architecture lightweight in the early training stage. The first stage is trained for approximately two days using the AdamW optimizer with a learning rate of $1 \times 10^{-5}$ on 64 A800 GPUs, with a batch size of 16 per GPU.

\textbf{Stage 2.} In the second stage, we train a fundamental MLLM integrated with a VPT encoder and multiple visual unit decoders. REF-VLM is trained on the constructed VT-Instruct dataset described in \Cref{sec:vt-instruct}, during which the parameters of all modules are jointly updated. The objective of this stage is to endow REF-VLM with the capability to predict visual primitives, such as bounding boxes and masks, under diverse interactive inputs, thereby supporting a wide range of downstream visual understanding and reasoning tasks. To this end, both the encoder and decoders of visual units are trained jointly, enabling effective cross-modal and cross-task feature alignment. The VPT encoder is randomly initialized and optimized together with the LLM and the visual unit decoders. A projector is employed to align the output dimension of the VPT encoder with the input dimension of the LLM, and its parameters are shared with the projector used between the visual encoder and the LLM to encourage representation consistency. For the visual unit decoders, we adopt a DETR-like architecture for the box decoder and a MaskFormer-like architecture for the mask decoder. All decoders are trained from scratch without pretrained weights, which helps maintain architectural coherence and avoids task-specific biases. This design further allows REF-VLM to rely on a unified visual encoder, simplifying deployment and reducing additional computational overhead. In this stage, we set the learning rate to $2 \times 10^{-6}$ and trained the model on A800 GPUs with a batch size of 16 per GPU for approximately nine days.

\textbf{Stage 3.} In the third stage, we train additional visual unit decoders based on our pretrained foundational MLLM from stage 2 to demonstrate the extensibility of our REF-VLM. We replaced the box decoder with GroundingDINO and the mask decoder with SAM. We then finetune the entire MLLM along with the newly integrated visual unit decoders separately. This demonstrated that our model could not only accommodate custom-designed decoders trained from scratch but also effectively leverage state-of-the-art (SOTA) visual decoders. For each separate training process for different visual decoders, We set the learning rate to $2e-5$ and trained the decoder for 5 epochs, using a batch size of 16 per GPU. 
\section{Experiments}
\label{sec:experiments}

We conduct comprehensive quantitative evaluations of REF-VLM across five tasks:(i) Visual Understanding, (ii) Grounded Conversation Generation (GCG), (iii) Referring Expression Comprehension, (iv) Freeform Open-Vocabulary Identification, and (v) Interactive Grounding. Furthermore, we perform ablation studies to assess the contribution of each key component in our approach.

\subsection{Quantitative Results}
\label{sec:quantitative_results}

\begin{table}[t]
\centering
\setlength{\abovecaptionskip}{3pt}
\setlength{\belowcaptionskip}{0pt}
\caption{Evaluation on Image Captioning and VQA for MLLMs.}
\resizebox{\columnwidth}{!}{%
\begin{tabular}{l|cc|cc}
    \toprule[1pt]
    \multirow{2}{*}{\textbf{Model}} &
    \multicolumn{2}{c|}{\textbf{Image Captioning}} &
    \multicolumn{2}{c}{\textbf{VQA}} \\ \cline{2-5}
     & Flickr30K & NoCaps & VQAv2 & OKVQA \\ \hline
    Flamingo-80B \cite{alayrac2022flamingo} & 67.2 & - & 56.3 & 50.6 \\
    InstructBLIP \cite{instructblip} & 82.8 & 121.9 & - & - \\
    LLaVA-1.5-7B \cite{llava} & - & - & 78.5 & 54.40 \\
    Shikra-13B \cite{chen2023shikra} & 73.9 & - & 77.4 & 47.16 \\
    InternVL-G \cite{chen2023internvl} & 79.2 & 113.7 & 80.2 & - \\
    Qwen-VL \cite{bai2023qwen} & 85.8 & 121.4 & 78.2 & - \\
    VisionLLMv2-Chat \cite{visionllmv2} & 88.7 & 118.1 & 81.4 & - \\
    VisionLLMv2 \cite{visionllmv2} & 90.2 & 116.9 & 80.8 & - \\
    GLaMM \cite{rasheed2024glamm} & 95.3 & 106.8 & - & - \\
    \rowcolor[HTML]{DAE8FC}
    REF-VLM & \textbf{96.0} & \textbf{122.4} & \textbf{81.6} & \textbf{62.39} \\
    \bottomrule[1pt]
\end{tabular}
}
\label{tab:visual_understanding}
\end{table}

\begin{table*}[t]
\centering
\setlength{\abovecaptionskip}{3pt}
\setlength{\belowcaptionskip}{0pt}
\caption{REG Comparison on RefCOCOg.}
\resizebox{\textwidth}{!}{%
\begin{tabular}{l|cccccccccc >{\columncolor[HTML]{DAE8FC}}c}
\toprule[1pt]
\textbf{Metric} 
& GRiT 
& Kosmos-2 
& ASM 
& RegionGPT 
& PixelLLM 
& GLaMM 
& Osprey 
& Groma 
& VisionLLMv2-Chat 
& VisionLLMv2 
& REF-VLM \\
\hline
CIDEr 
& 71.6 
& 62.3 
& 103.0 
& 109.9 
& 82.3 
& 106.0 
& 108.3 
& 107.3 
& 118.5 
& 111.6 
& \textbf{119.1} \\
Meteor 
& 15.2 
& 14.1 
& 20.8 
& 16.9 
& 14.3 
& 16.2 
& 16.6 
& 16.8 
& 21.2 
& 20.4 
& \textbf{21.6} \\
\bottomrule[1pt]
\end{tabular}
}
\label{tab:reg}
\end{table*}

\begin{table*}[!htbp]
    \centering
    \setlength{\abovecaptionskip}{3pt}
    \setlength{\belowcaptionskip}{0pt}
    \caption{Object hallucination benchmark in three POPE \cite{li2023evaluating} evaluation settings.}
    \label{tab:object_hallucination}
    \vspace{0.5em}
    \resizebox{\textwidth}{!}{%
    \begin{tabular}{ll
    ccccccccc >{\columncolor[HTML]{DAE8FC}}c}
    \toprule[1pt]
    Sampling & Metrics
    & GroundHOG \cite{zhang2024groundhog}
    & LION \cite{chen2024lion}
    & Osprey \cite{yuan2024osprey}
    & Ferret \cite{you2023ferret}
    & Shikra \cite{chen2023shikra}
    & \begin{tabular}[c]{@{}c@{}}LLaVA\\ -1.5 \cite{llava}\end{tabular}
    & \begin{tabular}[c]{@{}c@{}}Instruct\\ -BLIP \cite{instructblip}\end{tabular}
    & MiniGPT4 \cite{chen2023minigpt}
    & \begin{tabular}[c]{@{}c@{}}mPLUG\\ -Owl \cite{ye2023mplug}\end{tabular}
    & REF-VLM \\ \hline
    \multirow{4}{*}{Random} & Accuracy
    & 91.03 & 88.97 & 89.47 & 90.24 & 86.90 & 88.73 & 88.57 & 79.67 & 53.97 & \textbf{92.44} \\ 
     & Precision
    & 85.80 & \textbf{97.12} & 93.40 & 97.72 & 94.40 & 88.89 & 84.09 & 78.24 & 52.07 & 90.00 \\
     & Recall
    & 96.40 & 81.00 & 84.93 & 83.00 & 79.26 & 88.53 & 95.13 & 82.20 & \textbf{99.60} & 96.00 \\
     & F1 Score
    & 90.79 & 88.33 & 88.97 & 89.76 & 86.19 & 88.71 & 89.27 & 80.17 & 68.39 & \textbf{92.90} \\ \hline
    \multirow{4}{*}{Popular} & Accuracy
    & \textbf{90.13} & 86.77 & 87.83 & 84.90 & 83.97 & 85.83 & 82.77 & 69.73 & 50.90 & 90.10 \\
     & Precision
    & 85.93 & \textbf{91.69} & 89.94 & 88.24 & 87.55 & 83.91 & 76.27 & 65.86 & 50.46 & 85.78 \\
     & Recall
    & 93.81 & 80.87 & 85.20 & 80.53 & 79.20 & 88.67 & 95.13 & 81.93 & 99.40 & \textbf{96.13} \\
     & F1 Score
    & 89.70 & 85.94 & 87.50 & 84.21 & 83.16 & 86.22 & 84.66 & 73.02 & 66.94 & \textbf{90.66} \\ \hline
    \multirow{4}{*}{Adversarial} & Accuracy
    & 86.33 & 85.37 & 85.33 & 82.36 & 83.10 & 72.10 & 65.17 & 79.20 & 50.67 & \textbf{87.30} \\
     & Precision
    & 85.93 & 88.69 & 85.43 & 83.60 & 85.60 & 74.69 & 65.13 & 61.19 & 50.34 & \textbf{91.54} \\
     & Recall
    & 86.63 & 81.07 & 85.20 & 80.53 & 59.60 & 88.34 & \textbf{95.13} & 82.93 & 90.33 & 82.20 \\
     & F1 Score
    & 86.28 & 84.71 & 85.31 & 82.00 & 82.49 & 80.94 & 77.32 & 70.42 & 66.82 & \textbf{86.62} \\
    \bottomrule[1pt]
    \end{tabular}%
    }
\end{table*}

\noindent \textbf{Visual Understanding.} We begin by presenting quantitative comparisons on zero-shot image captioning tasks using the Flickr30k \cite{plummer2015flickr30k} and NoCaps \cite{agrawal2019nocaps} validation datasets, as well as VQA tasks on the VQAv2 \cite{VQA} and OK-VQA \cite{marino2019ok} test datasets. For image captioning, we report the CIDEr score, while for VQA tasks, overall accuracy is provided. The summarized results in \Cref{tab:visual_understanding} demonstrate that our REF-VLM model achieves the highest performance on the image captioning task, with a CIDEr score of 96.0 on the Flickr30k dataset and 122.4 on the NoCaps dataset. For VQA tasks, REF-VLM outperforms other models, achieving 62.39\% accuracy on the OK-VQA test dataset and 81.6\% on the VQAv2 test dataset, comparable to VisionLLMv2-Chat \cite{visionllmv2}. Furthermore, as shown in \Cref{tab:object_hallucination}, we utilize the POPE benchmark \cite{li2023evaluating} to assess the hallucination performance, where REF-VLM attains the highest F1 score, surpassing other MLLMs in every case.

\begin{table*}[!htbp]
\centering
 \setlength{\abovecaptionskip}{3pt} 
\setlength{\belowcaptionskip}{0pt} 
\caption{\textbf{REF-VLM performance on Grounding Conversation Generation (GCG) task.} Evaluation Metrics for GCG Tasks: CIDEr, Meteor, AP50, mIoU, and Recall. ``Scratch'' indicates whether the decoder is trained from scratch or utilizes a pretrained visual decoder (e.g., SAM \cite{kirillov2023segment}, Mask2Former \cite{cheng2021maskformer}). A \CheckmarkBold symbol indicates that the visual decoder in REF-VLM was trained from scratch, showcasing that REF-VLM outperforms models using pretrained decoders for generating boxes or masks.}
\label{tab:GCG}
\resizebox{\textwidth}{!}{%
\begin{tabular}{ccccccccc|ccccc}
\toprule[1pt]
 &  &  &  & \multicolumn{5}{c}{Val} & \multicolumn{5}{c}{Test} \\ \cline{5-14} 
\multirow{-2}{*}{Model} & \multirow{-2}{*}{Dataset} & \multirow{-2}{*}{Type} & \multirow{-2}{*}{Scratch} & CIDEr & Meteor & AP50 & mIoU & Recall & CIDEr & Meteor & AP50 & mIoU & Recall \\ \hline
BuboGPT \cite{zhao2023bubogpt} &  & Mask & \usym{2718} & 3.6 & 17.2 & 19.1 & 54.0 & 29.4 & 3.5 & 17.1 & 17.3 & 54.1 & 27.0 \\
Kosmos-2 \cite{peng2023kosmos} &  & Mask & \usym{2718} & 27.6 & 16.1 & 17.1 & 55.6 & 28.3 & 27.2 & 15.8 & 17.2 & 56.8 & 29.0 \\
LISA \cite{lai2024lisa} &  & Mask & \usym{2718} & 33.9 & 13.0 & 25.2 & 62.0 & 36.3 & 32.2 & 12.9 & 24.8 & 61.7 & 35.5 \\
GLaMM \cite{rasheed2024glamm} & \multirow{-4}{*}{$\text{GranD}_{f}$} & Mask & \usym{2718} & 47.2 & 16.2 & \textbf{30.8} & \textbf{66.3} & 41.8 & 37.9 & 14.6 & 27.2 & 64.6 & 38.0 \\
\rowcolor[HTML]{DAE8FC} 
REF-VLM &  & Mask & \CheckmarkBold & \textbf{56.9} & \textbf{18.4} & 26.2 & 57.9 & \textbf{50.0} & \textbf{53.2} & \textbf{21.7} & \textbf{27.7} & 56.6 & \textbf{45.3} \\ \hline
\rowcolor[HTML]{DAE8FC} 
REF-VLM & Flickr30k & Box & \CheckmarkBold & - & - & - & - & - & \textbf{82.0} & \textbf{26.0} & \textbf{35.4} & \textbf{66.1} & \textbf{47.7} \\ 
\bottomrule[1pt]
\end{tabular}%
}
\end{table*}

\begin{figure}[t]
  \centering
  \includegraphics[width=\columnwidth]{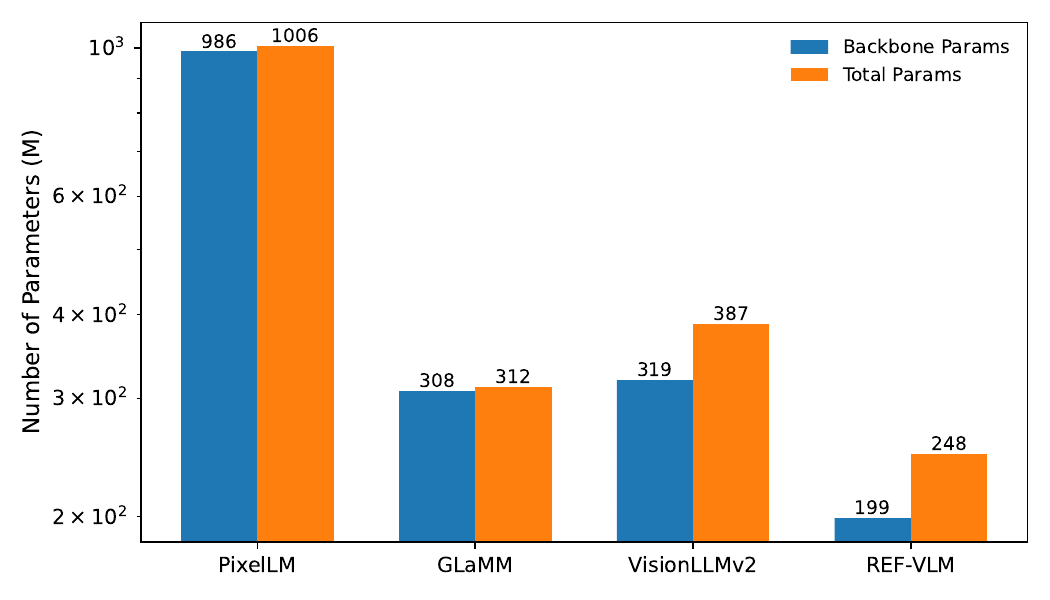}
  \caption{\textbf{The Comparison of parameter numbers.}
  We compared PixelLM \cite{xu2024pixel}, GLaMM \cite{rasheed2024glamm}, VisionLLMv2 \cite{visionllmv2}, and our REF-VLM in terms of the parameter count of the backbone used for feature extraction in the visual decoder module.}
  \label{fig:param_compare}
\end{figure}

\noindent \textbf{Grounded Conversation Generation (GCG).} The Grounded Conversation Generation (GCG) task consists of two components: GCG-mask and GCG-box. For the GCG-mask task, we further finetune our REF-VLM on the $\text{GranD}_f$ \cite{rasheed2024glamm} training dataset and evaluate its performance on the $\text{GranD}_f$ validation and test splits. The results presented in \Cref{tab:GCG} demonstrate that our REF-VLM, trained from the scratch outperforms current baseline methods which applied pretrained visual decoders, such as GLaMM \cite{rasheed2024glamm}, across metrics including CIDEr, Meteor, AP50, and Recall. Additionally, we assess the GCG-box task using the Flickr30k \cite{plummer2015flickr30k} test set, due to the lack of available MLLMs for the GCG-box task, we only report our zero-shot performance on this dataset.

\noindent \textbf{Referring Expression Generation (REG).} We evaluate Referring Expression Generation (REG) on the RefCOCOg test dataset \cite{kazemzadeh2014referitgame}, using Meteor and CIDEr as evaluation metrics. The results, shown in \Cref{tab:reg}, indicate that our REF-VLM outperforms the State-of-Art MLLM VisionLLMv2-Chat \cite{visionllmv2}. 

\noindent \textbf{Referring Expression Comprehension (REC).} We compare our REF-VLM with current MLLMs capable of generating referring boxes based on specific prompts in both zero-shot and fine-tuned settings. The metric used for REC evaluation is IoU@0.5. As shown in \Cref{tab:rec}, REF-VLM demonstrates superior performance in the REC task compared to other MLLMs. 

\noindent \textbf{Referring Expression Segmentation (RES).} For the Referring Expression Segmentation (RES) task, we evaluate REF-VLM on the RefCOCO, RefCOCO+, and RefCOCOg test and validation datasets by calculating the cumulative IOU (cIOU) as proposed by \cite{GRES}. REF-VLM with meta visual unit decoders, trained from scratch, achieves results in both zero-shot and fine-tuned settings that are comparable to recent methods like LISA \cite{lai2024lisa}, which utilized pretrained backbones such as SAM (see \Cref{tab:res}). The results show that the performance of our REF-VLM trained from scratch is slightly lower than that of PixelLM \cite{ren2024pixellm}, primarily due to PixelLM's use of the CLIP-ViT-H visual encoder, which has significantly more parameters than ours (see \Cref{fig:param_compare}). Additionally, to demonstrate that REF-VLM is not only capable of using custom-designed components but also can extend to current VGMs, we employed SAM as an external decoder for our mask decoder. By loading the pretrained weights from SAM and fine-tuning it on our VT-Instruct datasets, we found that REF-VLM with the external decoder outperforms current MLLMs such as VisionLLMv2 \cite{visionllmv2}, GLaMM \cite{rasheed2024glamm}, and demonstrates comparable performance to the Generalist Model such as UNINEXT-H \cite{yan2023universal}.

\noindent \textbf{Interactive Grounding} For this task, we evaluate using the prompt, “\texttt{Please generate a mask based on the region <region> in the image <image>.}” where \texttt{<region>} is replaced with visual prompts such as points, scribbles, boxes, or masks. The results presented in \Cref{tab:interactive_grounding} show that our REF-VLM with meta decoders outperforms both SAM \cite{kirillov2023segment} and SEEM-B \cite{zou2024seem} across point, scribble, box, and mask settings, achieving performance comparable to PSALM, which utilizes pretrained Swin-T and Mask2Former weights in these configurations. Furthermore, our REF-VLM with external decoders achieves superior performance compared to VisionLLMv2 \cite{visionllmv2} and PSALM \cite{zhang2024psalm}.

\begin{table*}[!htbp]
\centering
\caption{\textbf{Comparison of Referring Expression Comprehension (REC) performance.} REC is evaluated by IOU@0.5 Accuracy. VGM represents vision generalist model and MLLM represents multimodal large language model.}
\label{tab:rec}
\begin{tabular}{l|c|ccc|ccc|cc}
\toprule[1pt]
\multirow{2}{*}{Model} & \multirow{2}{*}{Type} & \multicolumn{3}{c|}{RefCOCO} & \multicolumn{3}{c|}{RefCOCO+} & \multicolumn{2}{c}{RefCOCOg} \\ \cline{3-10} 
 &  & Test-A & Test-B & Val & Test-A & Test-B & Val & Test & Val \\ \hline
OFA-L \cite{wang2022ofa}& \multirow{7}{*}{VGM} & 83.7 & 76.4 & 76.4 & 76.0 & 61.8 & 68.3 & 67.6 & 80.0 \\
MAttNet \cite{yu2018mattnet} &  & 80.4 & 69.3 & 80.0 & 70.3 & 56.0 & 64.9 & 67.0 & 76.4 \\
UNITER \cite{chen2020uniter} &  & 87.0 & 74.2 & 81.4 & 81.5 & 66.7 & 75.9 & 68.7 & 74.0 \\
VILLA \cite{gan2020large} &  & 87.5 & 74.8 & 82.4 & 81.5 & 66,8 & 76.2 & 76.7 & 76.2 \\
MDETR \cite{kamath2021mdetr} &  & 89.6 & 81.4 & 86.8 & 84.1 & 70.6 & 79.5 & 80.9 & 81.6 \\
GroundingDINO-T \cite{liu2023grounding}&  & 91.9 & 86.0 & 89.2 & 87.4 & 74.7 & 81.1 & 84.9 & 85.2 \\
GroundingDINO-L \cite{liu2023grounding}&  & 93.2 & 88.2 & 90.6 & 89.0 & 75.9 & 82.8 & 87.0 & 86.1 \\ \hline
Kosmos-2 \cite{peng2023kosmos}& \multirow{10}{*}{MLLM} & 57.4 & 47.3 & 52.3 & 50.7 & 42.2 & 45.5 & 61.7 & 60.6 \\
Shikra \cite{chen2023shikra}&  & 90.6 & 80.2 & 87.0 & 87.4 & 72.1 & 81.6 & 82.2 & 82.3 \\
Ferret \cite{you2023ferret}&  & 91.4 & 82.5 & 87.5 & 87.4 & 73.1 & 80.8 & 84.8 & 83.9 \\
NeXT-Chat \cite{zhang2023next}&  & 90.0 & 77.9 & 85.5 & 84.5 & 68.0 & 77.2 & 79.8 & 80.1 \\
MiniGPTv2-7B \cite{chen2023minigpt}&  & 91.3 & 84.3 & 88.1 & 85.5 & 73.3 & 79.6 & 84.3 & 84.2 \\
Qwen-VL-7B \cite{bai2023qwen}&  & 92.3 & 84.5 & 88.6 & 88.6 & 76.8 & 82.8 & 86.3 & 86.0 \\
VistaLLM \cite{pramanick2024jack}&  & 91.5 & 83.0 & 88.1 & 89.8 & 74.8 & 82.9 & 84.4 & 83.6 \\
VisionLLMv2 \cite{visionllmv2}&  & 93.1 & 87.1 & 90.0 & 87.3 & 74.5 & 81.1 & 84.8 & 83.9 \\
LION-12B \cite{chen2024lion}&  & 93.0 & 85.6 & 89.8 & 89.2 & 78.1 & 84.0 & 85.7 & 85.5 \\
\rowcolor[HTML]{DAE8FC} 
REF-VLM &  & \textbf{93.7} & \textbf{89.1} & \textbf{90.8} & 88.3 & \textbf{77.6} & 81.7 & \textbf{87.1} & \textbf{87.0} \\
\bottomrule[1pt]
\end{tabular}%
\end{table*}

\begin{table*}[!htbp]
\centering
\caption{\textbf{Comparison of interactive grounding performance on segmentation task.} The task is evaluated on the COCO-Interactive \cite{zhang2024psalm} validation dataset. The evaluation metrics for interactive grounding are mIoU and cIoU.}
\label{tab:interactive_grounding}
\resizebox{\textwidth}{!}{%
\begin{tabular}{l|c|c|c|cccccccc}
\toprule[1pt]
\multirow{2}{*}{Model} & \multirow{2}{*}{Decoder} & \multirow{2}{*}{Scratch} & \multirow{2}{*}{Type} & \multicolumn{2}{c}{Point} & \multicolumn{2}{c}{Scribble} & \multicolumn{2}{c}{Box} & \multicolumn{2}{c}{Mask} \\
 &  &  &  & mIoU & cIoU & mIoU & cIoU & mIoU & cIoU & mIoU & cIoU \\ \hline
SAM-B \cite{kirillov2023segment}& - & \CheckmarkBold & \multirow{3}{*}{VGM} & 48.7 & 33.6 & - & - & 73.7 & 68.7 & - & - \\
SAM-L \cite{kirillov2023segment}& - & \CheckmarkBold &  & 51.8 & 37.7 & - & - & 76.6 & 71.6 & - & - \\
SEEM-B \cite{zou2024seem}& - & \CheckmarkBold &  & 47.8 & 57.8 & 43.0 & 44.0 & 44.9 & 42.1 & 48.4 & 65.0 \\ \hline
PSALM \cite{zhang2024psalm}& Mask2Former & \usym{2718} & \multirow{2}{*}{MLLM} & 64.3 & 74.0 & 66.9 & 80.0 & 67.3 & 80.9 & 67.6 & 82.4 \\
VisionLLMv2 \cite{visionllmv2}& GroundingDINO & \usym{2718} &  & 65.4 & 70.9 & 66.8 & 77.2 & 74.2 & 83.2 & 67.9 & 83.8 \\
\rowcolor[HTML]{DAE8FC}
REF-VLM (meta) & Mask Decoder & \CheckmarkBold &  & 62.8 & 70.4 & 59.8 & 60.2 & 71.2 & 73.7 & 66.3 & 77.5 \\
\rowcolor[HTML]{DAE8FC}
REF-VLM (external) & SAM & \usym{2718} &  & \textbf{65.6} & \textbf{75.2} & \textbf{68.3} & 79.4 & \textbf{74.9} & \textbf{84.6} & \textbf{68.2} & 83.7 \\ 
\bottomrule[1pt]
\end{tabular}%
}
\end{table*}

\begin{table}[!htbp]
    \centering
     \setlength{\abovecaptionskip}{3pt} 
\setlength{\belowcaptionskip}{0pt} 
    \caption{\textbf{Evaluation on zero-shot open-vocabulary tasks.} Results on ADE20K (instance segmentation) and COCO2017 (object detection). As no existing MLLM supports zero-shot detection without class prompts, only the proposed model is evaluated.}
    \label{tab:open_vocab}
    \resizebox{\columnwidth}{!}{%
    \begin{tabular}{ccccccc}
    \toprule[1pt]
    Model & Type & Scratch & \begin{tabular}[c]{@{}c@{}}Mask\\ Decoder\end{tabular} & \begin{tabular}[c]{@{}c@{}}Box\\ Decoder\end{tabular} & ADE20k & COCO \\ \hline
    MaskCLIP \cite{dong2023maskclip} & SEG & \CheckmarkBold & MaskCLIP & - & 6.0 & - \\
    ODISE \cite{xu2023odise} & SEG & \CheckmarkBold & Diffusion UNet & - & 14.4 & - \\
    SAN \cite{xu2023san} & SEG & \CheckmarkBold & CLIP+SAN & - & 10.6 & - \\
    PSALM \cite{zhang2024psalm} & SEG & \usym{2718} & Mask2Former & - & 9.0 & - \\
    PSALM+LVIS \cite{zhang2024psalm} & SEG & \usym{2718} & Mask2Former & - & 13.9 & - \\
    \rowcolor[HTML]{DAE8FC} REF-VLM ($\text{mAP}_S$) & SEG/DET & \CheckmarkBold & MaskFormer & DETR & \textbf{16.7} & \textbf{26.7} \\ 
    \bottomrule[1pt]
    \end{tabular}%
    }
\end{table}

\begin{table*}[!htbp]
\centering
\caption{\textbf{Comparison of Referring Expression Segmentation (RES) performance.} The ``Decoder" column refers to the visual decoder utilized by the MLLM for performing RES tasks. $^*$ indicates that the visual decoder is custom-designed and trained from scratch. The performance of RES is evaluated using cumulative IoU (cIoU) as proposed by \cite{GRES}.}
\label{tab:res}
\resizebox{\textwidth}{!}{%
\begin{tabular}{l|c|c|c|ccc|ccc|cc}
\toprule[1pt]
\multirow{2}{*}{Model} & \multirow{2}{*}{Decoder} & \multirow{2}{*}{Sractch} & \multirow{2}{*}{Type} & \multicolumn{3}{c|}{RefCOCO} & \multicolumn{3}{c|}{RefCOCO+} & \multicolumn{2}{c}{RefCOCOg} \\ \cline{5-12} 
 &  &  &  & Test-A & Test-B & Val & Test-A & Test-B & Val & Test & Val \\ \hline
MCN \cite{mcn} & - & \CheckmarkBold & \multirow{7}{*}{VGM} & 64.2 & 59.7 & 62.4 & 55.0 & 44.7 & 50.6 & 49.4 & 49.2 \\
VLT \cite{vlt}& - & \CheckmarkBold &  & 70.5 & 65.2 & 67.5 & 61.0 & 50.1 & 56.3 & 57.7 & 55.0 \\
RELA \cite{GRES}& - & \CheckmarkBold &  & 76.5 & 70.2 & 73.8 & 71.0 & 57.7 & 66.0 & 66.0 & 65.0 \\
X-Decoder \cite{zou2022xdecoder}& - & \CheckmarkBold &  & - & - & - & - & - & - & - & 64.6 \\
SEEM-L \cite{zou2024seem}& - & \CheckmarkBold &  & - & - & - & - & - & - & - & 65.7 \\
UNINEXT-H \cite{yan2023universal}& - & \CheckmarkBold &  & 83.4 & 81.3 & 82.2 & 76.4 & 66.2 & 72.5 & 76.4 & 74.7 \\
GLEE-Pro \cite{wu2024general}& - & \CheckmarkBold &  & - & - & 80.0 & - & - & 69.6 & - & 72.9 \\ \hline
LISA-7B \cite{lai2024lisa}& SAM & \usym{2718} & \multirow{11}{*}{MLLM} & 72.3 & 79.1 & 74.9 & 70.8 & 58.1 & 65.1 & 70.6 & 67.9 \\
PixelLM \cite{ren2024pixellm}& Mask Decoder$^*$ & \CheckmarkBold &  & 76.5 & 68.2 & 73.0 & 71.7 & 58.3 & 66.3 & 70.5 & 69.3 \\
PixelLLM \cite{xu2024pixel}& SAM & \usym{2718} &  & 78.5 & 74.4 & 76.9 & 72.1 & 64.5 & 69.2 & 72.4 & 70.7 \\
AnyRef & SAM & \usym{2718} &  & 79.9 & 74.2 & 76.9 & 73.5 & 61.8 & 70.3 & 70.7 & 70.0 \\
NExT-Chat \cite{zhang2023next}& SAM & \usym{2718} &  & 78.9 & 69.5 & 74.7 & 71.9 & 56.7 & 65.1 & 67.0 & 67.0 \\
VITRON \cite{fei2024vitron}& SEEM & \usym{2718} &  & 78.7 & 71.6 & 74.4 & 72.1 & 57.8 & 66.3 & 67.3 & 67.2 \\
GroundHOG \cite{zhang2024groundhog}& Mask2Former & \usym{2718} &  & 79.9 & 75.7 & 78.5 & 75.0 & 64.9 & 70.5 & 74.6 & 74.1 \\
GLaMM \cite{rasheed2024glamm}& SAM & \usym{2718} &  & 83.2 & 76.9 & 79.5 & 78.7 & 64.6 & 72.6 & 74.9 & 74.2 \\
VisionLLMv2 \cite{visionllmv2}& GroundingDINO & \usym{2718} &  & 82.3 & 77.0 & 79.2 & 75.8 & 61.8 & 68.9 & 74.8 & 73.3 \\
\rowcolor[HTML]{DAE8FC}
REF-VLM (meta) & Mask Decoder$^*$ & \CheckmarkBold &  & 73.4 & 63.9 & 69.0 & 70.8 & 56.2 & 62.3 & 65.8 & 65.0 \\
\rowcolor[HTML]{DAE8FC}
REF-VLM (external) & SAM & \usym{2718} &  & 82.9 & 76.8 & 81.2 & 77.6 & 63.4 & 73.1 & 75.0 & 74.6 \\ 
\bottomrule[1pt]
\end{tabular}%
}
\end{table*}

\begin{table}[!htbp]
\centering
 \setlength{\abovecaptionskip}{1pt} 
\setlength{\belowcaptionskip}{0pt} 
\caption{\textbf{Comparison across visual encoder and ablation study on group matchers.} [0,1,2,3]: all CLIP-ConvNeXt layers; -2: second-to-last CLIP-ViT layer; [0,1,2,4]: first three CLIP-ConvNeXt layers plus CLIP-ViT output. \CheckmarkBold indicates the use of the Group Hungarian Matcher during decoder training.}

\label{tab:ablation_1}
\resizebox{\columnwidth}{!}{%
\begin{tabular}{c|cccc}
\toprule[1pt]
Visual Encoder & Size & \begin{tabular}[c]{@{}c@{}}Group\\ Matcher\end{tabular} & \begin{tabular}[c]{@{}c@{}}Feature\\ Dimension\end{tabular} & cIoU \\ \hline
ConvNeXt-L & 320 & \CheckmarkBold & {[}0,1,2,3{]} & 41.94 \\
ConvNeXt-L & 336 & \CheckmarkBold & {[}0,1,2,3{]} & 46.08 \\
CLIP-ViT & 336 & \CheckmarkBold & -2 & 60.44 \\
ConvNeXt-L + CLIP-ViT & 320 & \CheckmarkBold & {[}0,1,2,4{]} & 60.02 \\
\rowcolor[HTML]{DAE8FC} ConvNeXt-L + CLIP-ViT & 512 & \usym{2718} & {[}0,1,2,4{]} & 61.83 \\
\rowcolor[HTML]{DAE8FC} ConvNeXt-L + CLIP-ViT & 512 & \CheckmarkBold & {[}0,1,2,4{]} & \textbf{62.49} \\ 
\bottomrule[1pt]
\end{tabular}%
}
\end{table}


\noindent \textbf{Freeform Open-Vocabulary Identification.} Compared to existing MLLMs, which generally require category prompts to be added in open-vocabulary tasks \cite{visionllmv2,zhang2024psalm}, enabling the model to understand potential categories in the image before performing open-set detection or segmentation, our REF adopts a more flexible approach. With REF-VLM, there is no need to predefine categories, we simply input the prompt like “\texttt{Please detect bounding boxes (segment objects) in the image<image>}”. With this prompt, REF-VLM can autonomously identify potential categories within the image through its LLM, handing off decoding tasks to the subordinate visual decoder. We refer to this more adaptable task as \textit{Freeform Open-Vocabulary Identification}. Unlike other MLLMs \cite{visionllmv2,zhang2024psalm}, REF-VLM allows the LLM to directly output categories instead of having the downstream decoder output category names and confidence scores. Accordingly, instead of the calculating mAP as the metric for Open-Vocabulary Identification tasks, we propose a new metric called mAP Similarity ($\text{mAP}_S$) to evaluate our REF-VLM performance. For traditional open-vocabulary models, they typically predict classes with a logit score by their classification head. However, instead of applying a classification head for each task, our REF-VLM leverages a large language model (LLM) to predict classes without generating any class logits. We therefore compute the similarity score between REF-VLM’s class predictions and all ground truth class names. We then assign the class label based on the highest similarity, using this similarity score in place of the traditional confidence score. 

For the implementation of $\text{mAP}_S$, we denote the phrases predicted by the LLM as
$\mathcal{P} = \{p_1, p_2, \dots, p_k\}$,
where $k$ is the number of predicted phrases.
The ground-truth class set is defined as
$\mathcal{C} = \{c_1, c_2, \dots, c_n\}$,
where $n$ denotes the total number of classes in the dataset. We employ the CLIP-Large-14-336 model to encode both predicted phrases and ground-truth class names into a shared embedding space, yielding text embeddings
$\mathbf{e}_{p_i} = f_{\mathrm{CLIP}}(p_i)$ and $\mathbf{e}_{c_j} = f_{\mathrm{CLIP}}(c_j)$, respectively.
For each predicted phrase $p_i$, we compute the cosine similarity between its embedding and all class embeddings as $s_{i,j} =
\frac{
\mathbf{e}_{p_i}^{\top} \mathbf{e}_{c_j}
}{
\lVert \mathbf{e}_{p_i} \rVert_2
\lVert \mathbf{e}_{c_j} \rVert_2
}$. The predicted class for $p_i$ is then determined by
$j^{*} = \arg\max_{j} \; s_{i,j}$,
and the corresponding maximum similarity score $s_{i,j^{*}}$ is used as the logit score for evaluation.

We evaluate REF-VLM on the ADE20k test dataset for zero-shot freeform open-vocabulary segmentation task and COCO2017 validation dataset for object detection. The results for both tasks are presented in \Cref{tab:open_vocab}. Notably, REF-VLM achieves strong performance without any specialized design, outperforming other MLLMs (e.g., PSALM \cite{zhang2024psalm}) and specialist models (e.g., SAN \cite{xu2023san}). Additionally, REF-VLM also demonstrates the capability to perform open-vocabulary object detection.



\begin{figure*}[t]
  \centering
  \includegraphics[width=\textwidth]{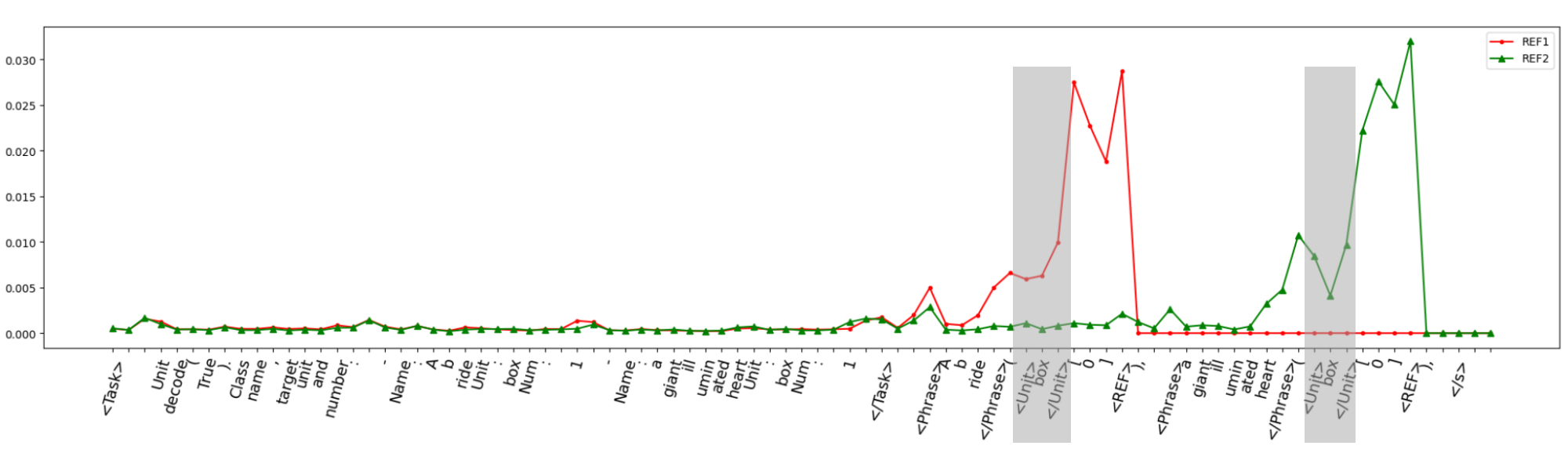}
  \caption{\textbf{Attention Significance of \textless REF\textgreater Tokens for MLLM Output Tokens.} Under the TRP paradigm, the generated prior information guides the formation of decoding features.}
  \label{fig:attention}
\end{figure*}

\noindent \textbf{Significance Analysis of TRP.} We selected a general segmentation task with the prompt ``\texttt{please segment objects in the image \textless image\textgreater.}" The model outputs two classification results: ``a bride" and ``a giant illuminated heart", using our designed VD-CoT format. We used attention visualization techniques to generate attention significance maps for the REF token with respect to each output token of the MLLM. The attention values from each layer were averaged and visualized, as shown in the figure above. The \Cref{fig:attention} demonstrates that each \textless REF\textgreater token exhibits high attention response values to the preceding \textless Phrase\textgreater, \textless Unit\textgreater, and output numbers.

\subsection{Ablation Study}
\label{sec:ablation_study}

\noindent \textbf{Choose of Group Matchers.} To validate the effectiveness of our Group Hungarian Matcher, we perform an ablation study on its usage in the mask decoder for the RES task, using the RefCOCOg test dataset and cIoU as the evaluation metric. As shown in \Cref{tab:ablation_1}, applying the Group Hungarian Matcher for loss computation yields a significantly better performance compared to configurations without it, demonstrating its substantial impact on improving the overall accuracy.

\noindent \textbf{Configuration of Visual Encoders.} To investigate the effect of different configurations of CLIP vision encoders, including CLIP-ConvNeXt and CLIP-ViT, along with variations in image size and feature selection layers, we conduct experiments on the RES task using the RefCOCOg test dataset. As shown in \Cref{tab:ablation_1}, REF-VLM achieves the highest performance when concatenating the CLIP-ConvNeXt and CLIP-ViT encoders with the setting of image size as 512×512.


\section{Conclusion}


In conclusion, we propose REF-VLM, an extensible open-ended visual multi-task learning framework, together with TRP, a unified referring paradigm, and VT-Instruct, a large-scale multimodal instruction-tuning dataset. Extensive experiments demonstrate the effectiveness of REF-VLM across diverse visual tasks and settings. Nonetheless, current task coverage remains limited, and performance in multi-turn dialogues as well as data utilization require further investigation. Future work will focus on expanding supported tasks, improving multi-turn robustness, and optimizing training data usage efficiency.


\bibliographystyle{IEEEtran}
\bibliography{main_1}
\clearpage

\begin{IEEEbiography}[{\includegraphics[width=1in,height=1.25in,clip,keepaspectratio]{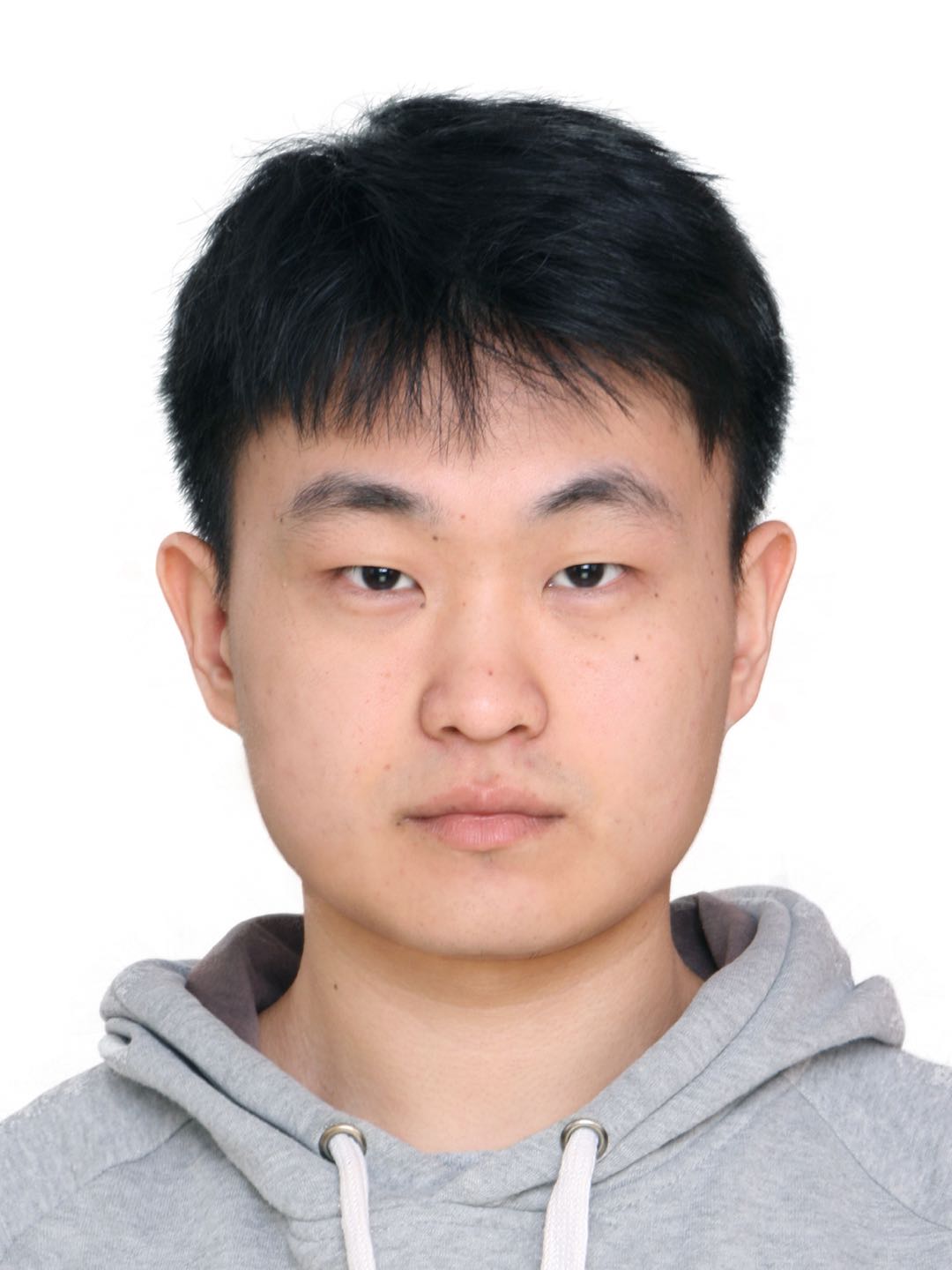}}]{Yan Tai}
received the M.S. degree from the Institute of Automation, Chinese Academy of Sciences, Beijing, China, in January 2024. He is currently pursuing the Ph.D. degree in computer science with Shanghai Jiao Tong University, Shanghai, China. His research interests include multimodal large language models, image generation, and embodied intelligence.
\end{IEEEbiography}

\begin{IEEEbiography}[{\includegraphics[width=1in,height=1.25in,clip,keepaspectratio]{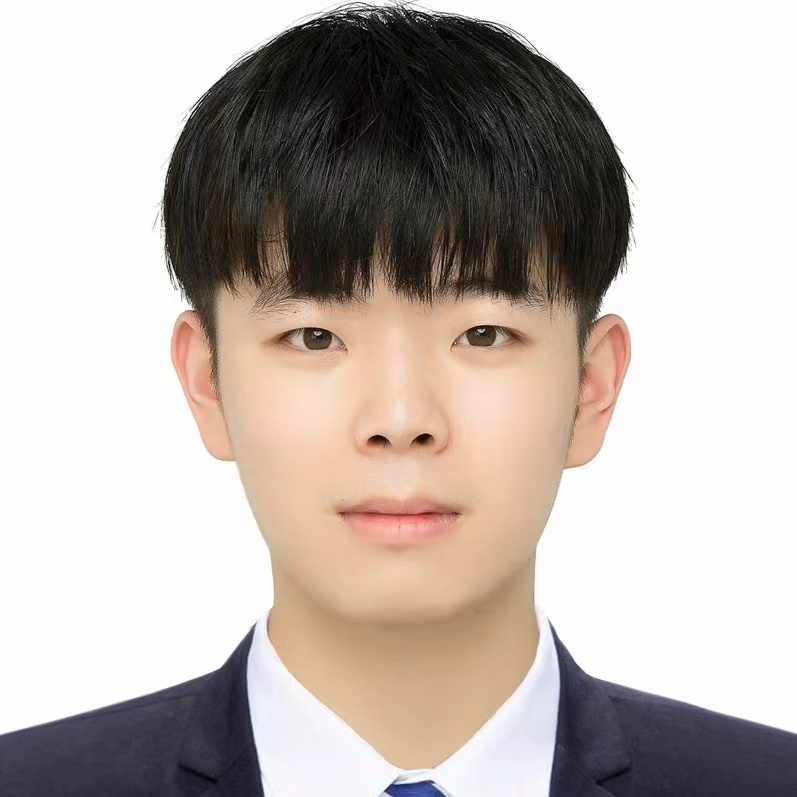}}]{Luhao Zhu}
received the B. Mgt. degree in human resource management from School of Management, Zhejiang University, China in 2021. Currently, he is a master student in management science and engineering at School of Management, Zhejiang University, China. His research interests include explainable artificial intelligence, adversarial learning, and generative model.
\end{IEEEbiography}

\begin{IEEEbiography}[{\includegraphics[width=1in,height=1.25in,clip,keepaspectratio]{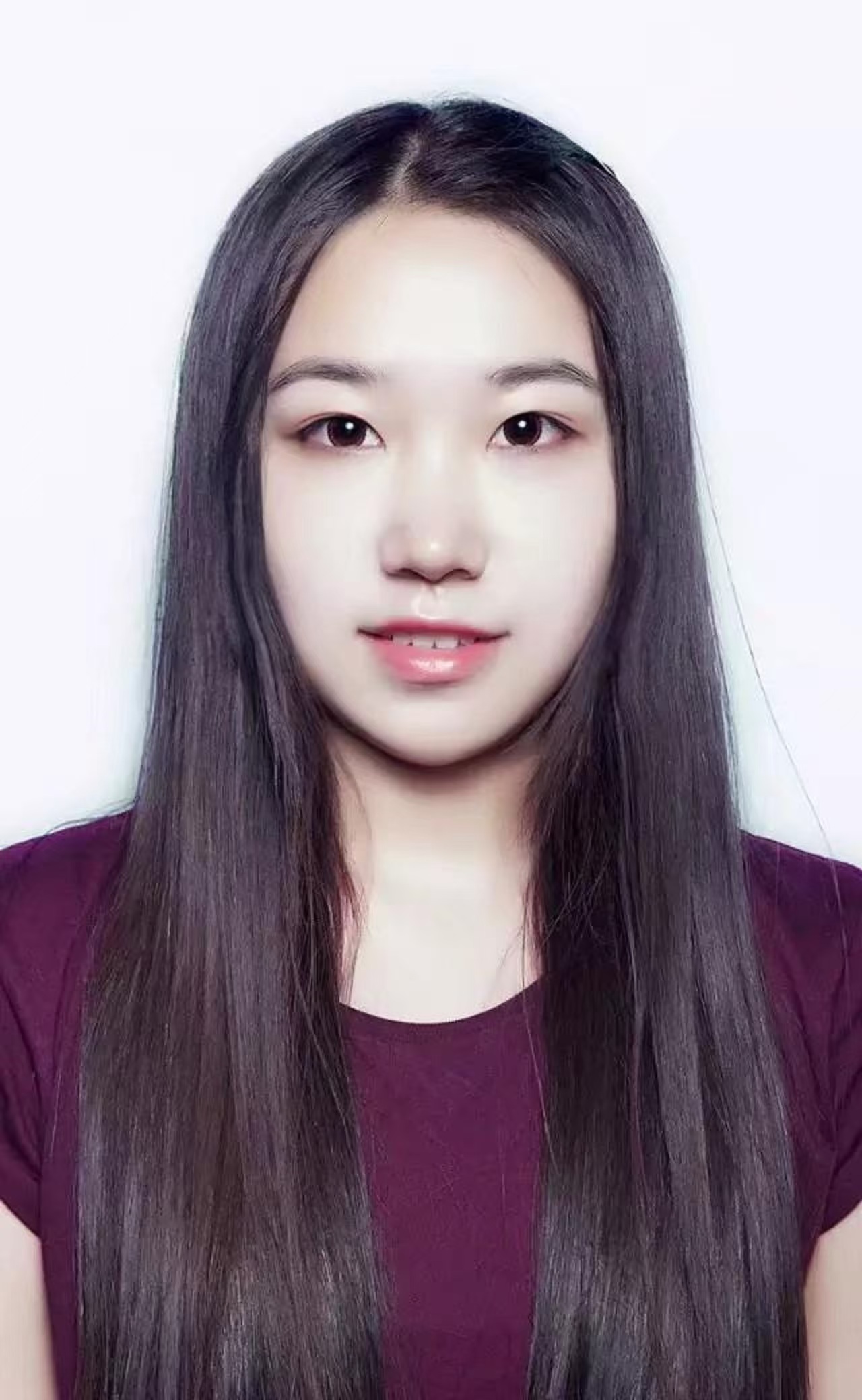}}]{Yunan Ding}
received the B.E. degree in Measurement and Control Technology and Instrumentation from Northeastern University, China in 2019, and the M.Sc. degree in Electronic Information Engineering from City University of Hong Kong. She is currently pursuing the Ph.D. degree with the Department of Computing, The Hong Kong Polytechnic University. Her research interests include multimodal large language models, model pruning and quantization.

\end{IEEEbiography}

\begin{IEEEbiography}[{\includegraphics[width=1in,height=1.25in,clip,keepaspectratio]{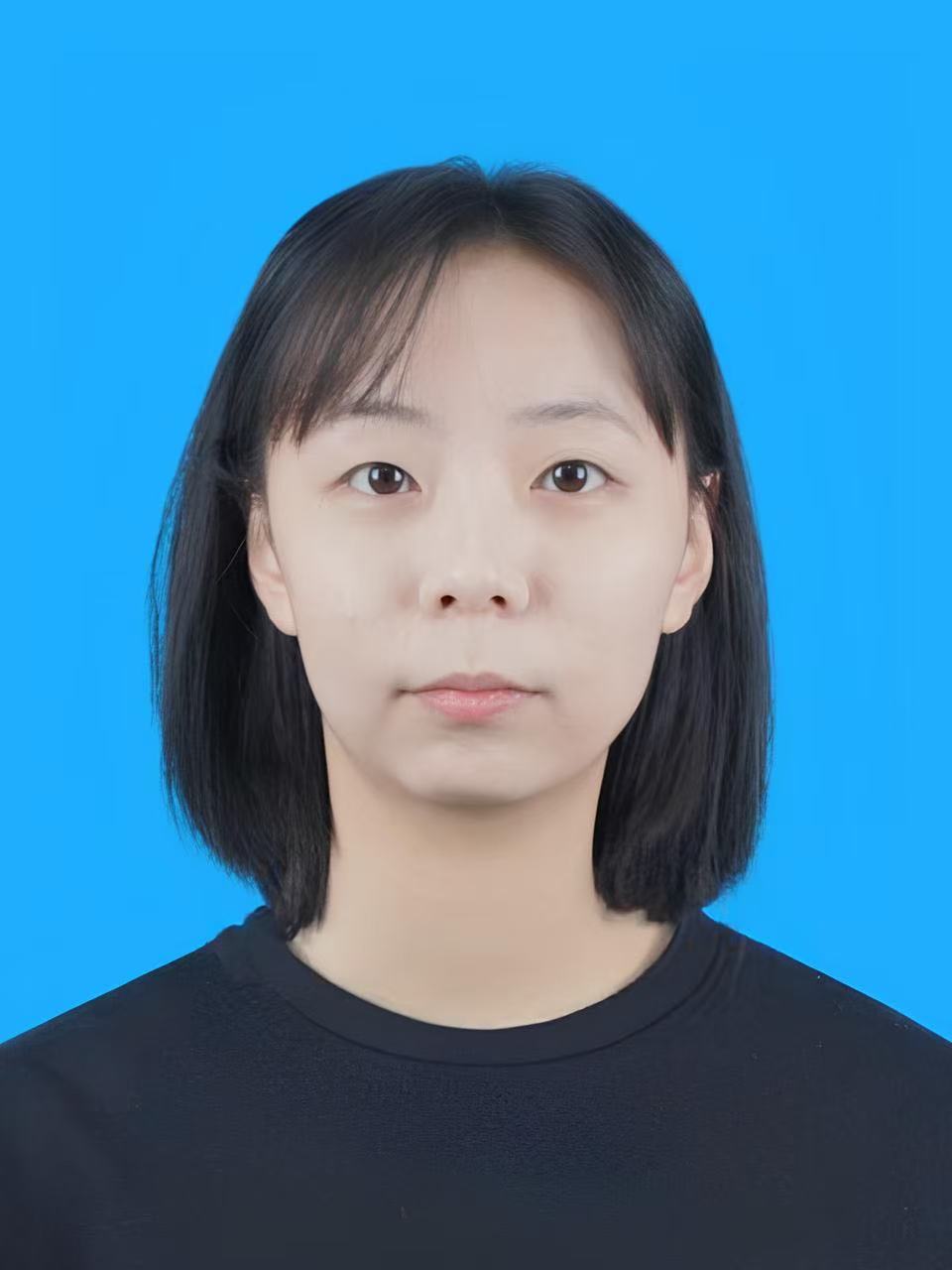}}]{Yiying Dong}
received her B.E. degree in computer science from Chongqing University of Posts and Telecommunications in 2022. She is currently pursuing a Ph.D. degree at the Department of Computing, The Hong Kong Polytechnic University, China. Her current research interests include multimodal reasoning and generative models for visual understanding.

\end{IEEEbiography}

\begin{IEEEbiography}[{\includegraphics[width=1in,height=1.25in,clip,keepaspectratio]{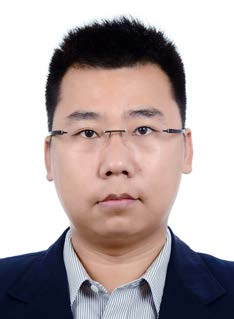}}]{Guangtao Zhai (IEEE Fellow)}
received the
B.E. and M.E. degrees from Shandong University, Shandong, China, in 2001 and 2004, respectively,
and the PhD degree from Shanghai Jiao Tong University, Shanghai, China, in 2009, where he is currently a Research Professor with the Institute of Image Communication and Information Processing.
From 2008 to 2009, he was a Visiting Student
with the Department of Electrical and Computer
Engineering, McMaster University, Hamilton, ON,
Canada, where he was a Post-Doctoral Fellow from
2010 to 2012. From 2012 to 2013, he was a Humboldt Research Fellow with
the Institute of Multimedia Communication and Signal Processing, Friedrich
Alexander University of Erlangen-Nuremberg, Germany. He received the
Award of National Excellent PhD Thesis from the Ministry of Education
of China in 2012. His research interests include multimedia signal processing
and perceptual signal processing.
\end{IEEEbiography}

\begin{IEEEbiography}[{\includegraphics[width=1in,height=1.25in,clip,keepaspectratio]{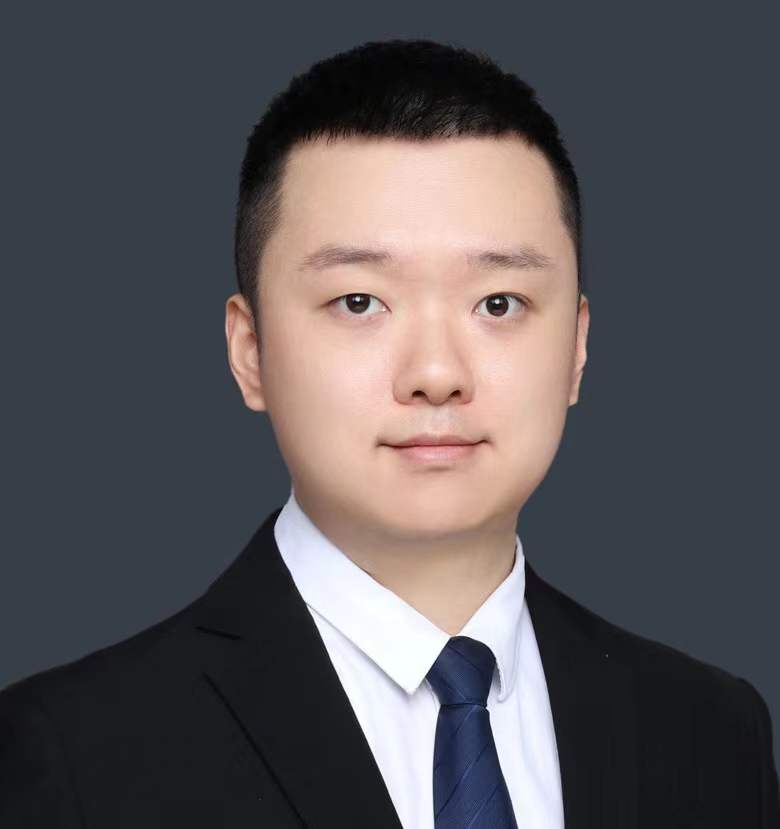}}]{Xiaohong Liu (Member, IEEE)}
received the PhD degree in elec-
trical and computer engineering from McMaster
University, Canada, in 2021. He is currently an
Associate Professor with the School of Electronic
Information and Electrical Engineering, Shanghai
Jiao Tong University. His research interests lie in
computer vision and multimedia. He has published
more than 80 academic papers in top journals and
conferences such as IEEE TIP, IEEE TMM, CVPR,
ICCV, ECCV, etc., and has been awarded the Mi-
crosoft Research Asia StarTrack Scholars Program
in 2024, Shanghai Pujiang Program in 2022, Chinese Government Award
for Outstanding Self-financed Students Abroad in 2021, and Borealis AI
Fellowships in 2020. His research has been supported by the Young Scientists
Fund of the National Natural Science Foundation of China (NSFC), Young
Scientists Fund of Natural Science Foundation of Sichuan Province, and
many high-tech companies. He also serves as Associate Editor of the ACM
Transactions on Multimedia Computing, Communications, and Applications
(TOMM).
\end{IEEEbiography}

\begin{IEEEbiography}[{\includegraphics[width=1in,height=1.25in,clip,keepaspectratio]{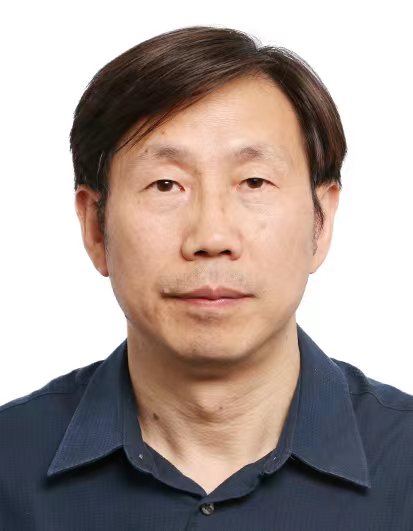}}]{Guodong Guo (Senior Member, IEEE)}
received the
BE degree in automation from Tsinghua University,
Beijing, China, and the PhD degree in computer science from the University of Wisconsin-Madison. He
is a professor with the Lane Department of Computer
Science and Electrical Engineering, West Virginia
University. In the past, he visited and worked in
several places, including INRIA, Sophia Antipolis,
France, Ritsumeikan University, Japan, Microsoft
Research, China, and North Carolina Central University. He won the North Carolina State Award for
Excellence in Innovation in 2008, and Outstanding New Researcher of the Year
(2010-2011) at CEMR, WVU. His research areas include computer vision,
machine learning, and multimedia. He has authored or coauthored 4 books,
published about 200 technical papers. He is an editorial board member of the
IET Biometrics, Journal of Visual Communication and Image Representation,
and IEEE Transactions on Affective Computing.

\end{IEEEbiography}

\vfill

\end{document}